\documentclass[journal]{IEEEtran}
\usepackage[utf8]{inputenc}
\usepackage{enumitem}
\usepackage{url}
\usepackage{graphics}
\usepackage{tikz}
\usepackage{caption}
\usepackage{subcaption}
\usepackage{multirow}
\usepackage{graphicx}
\usepackage{tabularx}
\usepackage{listings}
\usepackage{float}
\usepackage{amsmath}
\usepackage{amssymb}
\usepackage{mathtools}
\usepackage{booktabs}
\usepackage{graphicx, subcaption, adjustbox}
\usepackage{tcolorbox}
\tcbuselibrary{skins, breakable}
\usepackage{cite}
\usepackage{nowidow}
\usepackage{balance}
\newcolumntype{Y}{>{\RaggedRight\arraybackslash}X}

\usepackage[norelsize, linesnumbered, ruled, vlined,
boxed, commentsnumbered]{algorithm2e}

\SetKwInput{KwOutput}{Output}
\SetKwInput{KwInput}{Input}
\newcommand{\nosemic}{%
  \renewcommand{\@endalgocfline}{\relax}}
\newcommand{\dosemic}{%
  \renewcommand{\@endalgocfline}{\algocf@endline}}

\let\oldnl\nl
\newcommand{\nonl}{%
  \renewcommand{\nl}{\let\nl\oldnl}}

\setlist[itemize]{noitemsep, topsep=0pt, leftmargin=*}
\definecolor{mypurple}{RGB}{128,0,128}

\newcommand{\blue}[1]{#1}

\definecolor{macrogood}{RGB}{100,160,220}
\definecolor{macrobad}{RGB}{220,120,120}
\definecolor{corebg}{RGB}{245,245,245}
\definecolor{goodcheck}{RGB}{40,160,60}
\definecolor{badcross}{RGB}{200,40,40}
\definecolor{pincolor}{RGB}{255,180,0}
\definecolor{iocolor}{RGB}{180,80,200}
\definecolor{whitespaceok}{RGB}{200,235,200}
\definecolor{whitespacefrag}{RGB}{255,220,220}
\definecolor{pocketcolor}{RGB}{255,200,200}

\begin{document}

\title{MAGE: Human-Like Macro Placement via Agentic Multimodal Reasoning}

\author{
    Andrew B. Kahng,~\IEEEmembership{Fellow,~IEEE}, 
    Sayak Kundu,~\IEEEmembership{Graduate Student Member,~IEEE},
    Bodhisatta Pramanik,~\IEEEmembership{Graduate Student Member,~IEEE}
}

\maketitle

\begin{abstract}
Macro placement still requires substantial manual refinement in industrial
physical design flows. We present {\em MAGE} (\textbf{M}acro Placement
\textbf{Ag}entic \textbf{E}ngine), a multimodal multi-agent framework for
macro placement refinement. {\em MAGE} decomposes the macro placement task into a six-phase workflow that
combines structured floorplanning rules, visual checks, and iterative
refinement. Expert floorplanning knowledge is encoded through natural-language
directives and validation criteria, rather than learned from labeled placement
data. A tournament-style refinement mode evaluates multiple candidate
placements and propagates feedback from higher-quality solutions.
We also introduce four metrics for quantifying human-likeness in macro
placement: notch score, whitespace score, pocket score, and alignment score.
These metrics capture structural properties used by expert designers but not
directly measured by conventional PPA metrics. Across nine designs in NanGate45
and GlobalFoundries 12nm enablements, {\em MAGE} achieves geometric-mean
improvements of 11.1\%--19.3\% in WNS and 70.0\%--74.0\% in TNS over commercial
macro placers. On the three NanGate45 designs, for which human-expert and
Hier-RTLMP baselines are available, {\em MAGE} improves WNS and TNS by 18.3\%
and 72.5\% over the human expert, and by 47.0\% and 80.4\% over Hier-RTLMP,
with comparable wirelength and power. On human-likeness metrics, {\em MAGE}
improves the overall score by
6\%--48\% over all baselines. Additional case studies on anonymized netlists,
unseen designs, dense rectilinear floorplans, and high-utilization settings
show that the framework transfers to new placement settings without
design-specific retraining.
\end{abstract}

\section{Introduction}
\label{sec:introduction}

Macro placement is a central problem in VLSI physical design. The task is to
determine the locations and orientations of large pre-designed blocks, such as
SRAMs, register files, analog macros, and accelerators, within a chip
floorplan. Since macros occupy substantial area and constrain whitespace,
routing access, and power delivery, their placement \textcolor{black}{strongly affects}
downstream PPA and design convergence. The problem is NP-hard and \textcolor{black}{in practice} requires
\textcolor{black}{optimization of} coupled objectives \textcolor{black}{that include} wirelength, congestion, \textcolor{black}{density}, routability, and design-rule feasibility. \textcolor{black}{Today}, high-quality macro placement relies heavily on expert \textcolor{black}{physical design (PD)}
engineers \textcolor{black}{who} reason not only about physical cost metrics, but
also about logical hierarchy, dataflow, symmetry, pin accessibility,
and \textcolor{black}{more}. By grouping \textit{related} macros
and arranging them into regular spatial patterns, engineers try to produce floorplans
that are structured, interpretable, and more amenable to optimization in later
implementation stages.

Despite decades of research, existing macro placement methods
\cite{AutoDMP, Lu2015, Kim2012, Mirhoseini2020,
Kahng2022RTLMP, Mirhoseini2021, Kahng2024HierRTLMP}
do not consistently reproduce this form of human expert reasoning. Classical
optimization methods, analytical placers, and learning-based techniques can
produce legal placements with competitive metric values. However, they often
miss the higher-level spatial regularity and functional coherence that
characterize expert-designed floorplans. As a result, automated placements may
require substantial manual refinement before they are suitable for downstream
implementation. 
In industrial flows, this refinement is often performed by experienced
\textcolor{black}{PD engineers} after an initial macro placement is produced by a commercial tool. 
The designers modify the placement to better reflect
hierarchy, dataflow, alignment, routing access, power-grid constraints, and
overall floorplan intent. This motivates automated macro placement methods that can
incorporate expert floorplanning principles directly into the macro placement
process.

This work addresses the above macro placement refinement challenge by encoding expert
floorplanning principles into a multimodal, validation-driven placement loop. 
Rather than optimizing only
for legality or proxy cost objectives, we seek macro placements that better
capture the organizational principles used by expert \textcolor{black}{PD} engineers. We propose
{\em MAGE} (\textbf{M}acro Placement \textbf{Ag}entic \textbf{E}ngine), a
multi-agent framework for macro placement that uses multimodal reasoning to
generate human-like \textcolor{black}{macro placements}. Instead of training a model on labeled
placement examples, {\em MAGE} encodes expert floorplanning knowledge through
structured prompts, agent interactions, visual checks, and explicit refinement
criteria. Our main contributions are as follows.

\begin{itemize}

\item \textit{A multi-agent framework for human-like macro placement.}
We propose {\em MAGE}, a multi-agent framework that generates macro placements
through ten specialized agents organized into a six-phase pipeline
(Section~\ref{sec:approach}). The agents handle design interpretation, macro
grouping, spatial organization, feasibility checking, refinement, and visual
audit. This decomposition treats macro placement as a structured design
process rather than a single monolithic optimization task. To the best of our
knowledge, {\em MAGE} is the first \textcolor{black}{agentic} 
macro placement framework
explicitly designed to generate human-like macro placement solutions.


\item \textit{Two-stage synthesis and evolutionary refinement.}
{\em MAGE} decomposes macro placement into two stages. The first stage
generates placements that prioritize structural properties such as coherent
macro grouping, interpretable spatial organization, boundary alignment, and
regular macro stacks (Section~\ref{sec:approach}). The second stage applies
a Go-With-The-Winners (GWTW)~\cite{AldousV94} tournament-style evolutionary optimization to
improve implementation metrics,
including wirelength and timing, while preserving the structural properties
established in the first stage (Section~\ref{sec:tournament}). We further enable
the framework to explore multiple candidate macro placements rather than following a
single macro placement trajectory.

\item \textit{Principles and metrics for human-likeness in macro placement.}
We formalize a set of floorplanning principles that characterize human-like
macro placement (Section~\ref{sec:placement-rules}). 
These principles define the structural
properties that {\em MAGE} seeks to preserve during placement. 
{\em MAGE} encodes these principles
through structured prompts, refinement criteria, visual checks, and a
supporting knowledge corpus, rather than learning them from labeled placement
data. We further
introduce four metrics to quantify human-likeness: notch score, whitespace
score, pocket score, and alignment score
(Section~\ref{sec:human_likeliness_metrics}). These metrics capture
floorplan structure that is important to expert-designed placements but is
not directly measured by standard PPA metrics.

\item \textit{Evaluation on standard, industrial, and rectilinear floorplans.}
We evaluate {\em MAGE} on designs across NanGate45 and GlobalFoundries 12nm
enablements, including standard macro placement designs from the
MacroPlacement repository~\cite{TILOS-repo}, designs from
Hier-RTLMP~\cite{Kahng2022RTLMP}, four rectilinear floorplan variants, and two
unseen designs. To the best of our knowledge, this is the first macro
placement work to evaluate placements on rectilinear floorplans. We compare
against a human expert baseline, two releases of \textcolor{black}{an unnamed commercial} 
macro placer,
and OpenROAD's Hier-RTLMP~\cite{Kahng2022RTLMP}.

\item \textit{Improved PPA over baselines.} 
{\em MAGE} achieves geometric-mean improvements of 47.0\% in WNS and 80.4\%
in TNS over Hier-RTLMP, and 11.1\%--19.3\% in WNS and 70.0\%--74.0\% in TNS
over the commercial baselines (Section~\ref{sec:eval_ppa}). On dense
rectilinear designs with over 70\% utilization, {\em MAGE} achieves
5\%--40\% better WNS and up to 74\% better TNS than the commercial baseline
(Section~\ref{sec:case_studies}). On human-likeness metrics, {\em MAGE}
outperforms all baselines by 6\%--48\% in overall human-likeness score
(Section~\ref{sec:human_metrics}). On two unseen designs not used during
prompt development or knowledge-corpus construction, {\em MAGE} improves WNS
by 6\%--27\% and human-likeness by 26\%--40\% over the commercial baselines
(Section~\ref{sec:case_studies}).

\end{itemize}

Table~\ref{tab:notation} summarizes the notation and terminology used in this
paper. The remainder of the paper is organized as follows.
Section~\ref{sec:related-work} reviews related work.
Section~\ref{sec:placement-rules} codifies principles of human-like macro
placement, defines four human-likeness metrics, and formalizes the
optimization objective. Section~\ref{sec:approach} describes the {\em MAGE}
framework. Section~\ref{sec:vision-use} explains how multimodal reasoning and
visual feedback are integrated into the placement pipeline.
Section~\ref{sec:experiments-results} presents the experimental results, and
Section~\ref{sec:conclusion} concludes the paper.

\begin{table}[t]
\centering
\caption{Summary of notation and terminology.}
\label{tab:notation}
\scriptsize
\setlength{\tabcolsep}{4pt}
\renewcommand{\arraystretch}{1.12}
\begin{tabular}{@{}c >{\raggedright\arraybackslash}p{0.62\columnwidth}@{}}
\toprule
\textbf{Symbol} & \textbf{Description} \\
\midrule
$\mathcal{N},\ \mathcal{F},\ \Theta$ & Design netlist, floorplan spec., configuration \\
$\mathcal{M}=\{m_1,\ldots,m_n\}$ & Set of hard macros \\
$(w_i,h_i),\ (x_i,y_i)$ & Dimensions and coordinates of macro $m_i$ \\
$o_i\in\mathcal{O}$ & Orientation of macro $m_i$ \\
$\mathcal{S},\ \mathcal{C},\ \mathcal{I}$ & Standard cells, core region, IO/pin info \\
$\mathcal{K}$ & Knowledge corpus \\
$\mathcal{X}^{0},\ \mathcal{X}^{\star}$ & Initial and refined macro placement \\
\midrule
$\mathcal{G}=\{g_1,\ldots,g_k\}$ & Set of hierarchy-aware macro groups \\
$\mathcal{Q}_1,\ldots,\mathcal{Q}_p$ & Families of similar macro groups \\
$B_g,\ W_g,\ H_g$ & Bounding box of group $g$, with width and height \\
$g_{\mathrm{ref}},\ \mathcal{P}_{\mathrm{ref}}$ & Reference group and its macro placement \\
$\mathcal{P},\ \mathcal{P}^{\star}$ & Chip-level and final selected macro placement \\
$T_g$ & Geometric transformation applied to group $g$ \\
$\tau_{\max}$ & Group-size threshold for recursive sub-grouping \\
\midrule
$N_v,\ N_s,\ R$ & Tournament variants, survivors, rounds (default $6,3,6$) \\
$\mathcal{V}^{(r)},\ \mathcal{W}^{(r)}$ & Variant pool and survivor set at round $r$ \\
$\mathcal{H}$ & Evaluation history (placements and eGR wirelengths) \\
$N_{\max}$ & Max.\ validation iterations (default $3$) \\
\midrule
$c_w$ & Minimum channel width between macros \\
$S_{\mathrm{notch}},\,S_{\mathrm{ws}},\,S_{\mathrm{pocket}},\,S_{\mathrm{align}}$ & Notch, whitespace, pocket, alignment scores \\
$S_{\mathrm{HM}}$ & Overall human-likeness score \\
\bottomrule
\end{tabular}
\end{table}

\section{Related Work}
\label{sec:related-work}

Macro placement has been extensively studied in the literature. We group prior
work into four categories: classical floorplanning, analytical and mixed-size
placement, hierarchy-aware macro placement, and learning-based methods.

\noindent \textbf{Classical macro placement.}
Early macro placement methods are largely based on simulated annealing
(SA)~\cite{Kirkpatrick1983} and compact geometric representations. \cite{Murata1996} introduced
the sequence-pair representation, and \cite{Chang2000} proposed B$^*$-trees
for nonslicing floorplans. \cite{AdyaMarkov2003} extended these ideas to
fixed-outline floorplanning for hierarchical designs. Subsequent works incorporated
additional objectives, including interconnect-driven optimization~\cite{Chen2008,Cong2006},
routability awareness~\cite{Chen2014,Chiou2016,Lin2019TVLSI}, and dataflow-aware
floorplanning~\cite{Lin2021}. \cite{Yan2008} further proposed
deferred-decision techniques to improve search quality. 


\noindent \textbf{Analytical and mixed-size placement.}
Analytical placers formulate placement as a continuous optimization problem and have been
widely used for mixed-size designs~\cite{AutoDMP}. 
\cite{Lu2015} presented {\em ePlace-MS}, an
electrostatics-based mixed-size placer. ~\cite{ChengKKW18} later extended this
in {\em RePlAce}, which introduced additional improvements. 
\cite{Kim2012} proposed {\em MAPLE}, a multilevel adaptive
placement framework, and \cite{Hsu2014} developed {\em NTUplace4h} for
hierarchical mixed-size placement. \cite{Yan2014} proposed a floorplan-guided
approach for large-scale mixed-size placement. These methods integrate macro placement into the 
standard-cell placement framework. 

\noindent \textbf{Hierarchy-aware macro placement.}
Previous works have incorporated hierarchy and architectural structure into placement~\cite{Hu2004}.
\cite{Chuang2010} \textcolor{black}{uses} design hierarchy to improve routability in mixed-size
placement, while \cite{KimLim2008, Ekpanyapong2006, Nookala2005} \textcolor{black}{addresses} microarchitectural
floorplanning. \cite{ChoiBazargan2003} \textcolor{black}{combines} SA with
network-flow-based area migration in a hierarchical framework. 
RTL-MP~\cite{Kahng2022RTLMP} \textcolor{black}{leverages} RTL hierarchy to 
generate human-quality placements
through a top-down partitioning and bottom-up macro placement methodology.
Hier-RTLMP~\cite{Kahng2024HierRTLMP} extends this approach to larger and more complex IP
blocks through hierarchical clustering. 


\noindent \textbf{Learning-based approaches.}
Machine learning (ML) has also been explored for macro placement. Early work
by \cite{Gwee1999} \textcolor{black}{has} applied genetic algorithms, and \cite{He2020}
learned local-search heuristics for placement improvement. More recently,
\cite{Mirhoseini2020,Mirhoseini2021} proposed deep reinforcement
learning for chip floorplanning. However, reinforcement-learning-based approaches often
require expensive training, can be sensitive to design distribution, and may be difficult
to adapt to industrial design constraints~\cite{ChengKKWW23}.

Across these prior methods, differences mainly arise from optimization
formulations, geometric representations, hierarchy modeling, or learning
strategy. Most closely related to our work, {\em VeoPlace}~\cite{UchenduGHSLJ26}
uses a \textcolor{black}{vision language model} (VLM) to guide existing 
low-level placers by proposing placement regions
or soft anchors, with HPWL improvement as the primary objective under an
inference-time rollout budget. 
\textcolor{black}{In contrast,
{\em MAGE} is not a wrapper around a single base placer; rather, it is a
multi-agent macro placement framework that uses multimodal reasoning,
structured expert knowledge, visual feedback, and explicit refinement criteria
to make placement decisions directly. Thus, while prior LLM-based EDA
efforts~\cite{Liu2024ChipNeMo, He2024ChatEDA, Thakur2024VeriGen, GhoseJKL26}
primarily use foundation models for code generation, verification,
documentation, design assistance, or tool scripting, {\em MAGE} applies
multimodal foundation models to the physical-design task itself, targeting
human-like structural organization in addition to conventional PPA objectives.}

\begin{figure}[htpb]
\centering
  \includegraphics[width=0.9\columnwidth]{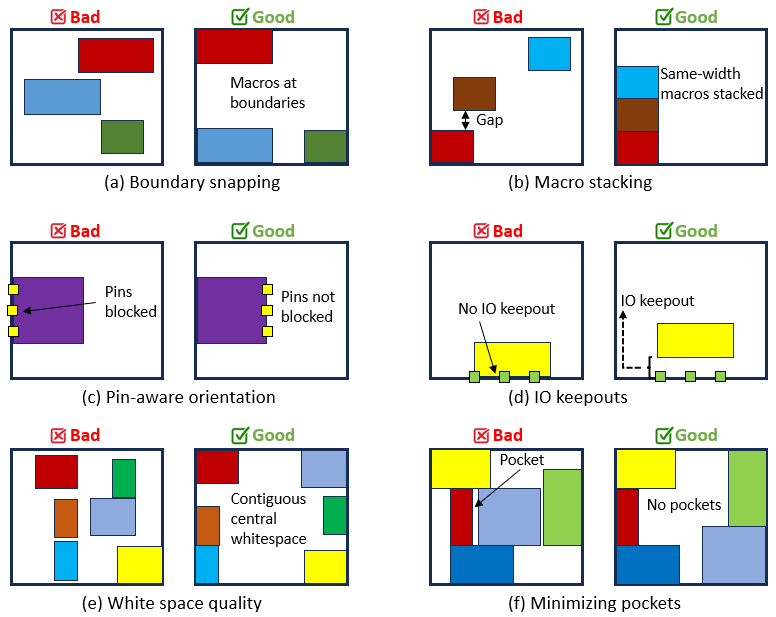}
  \caption{Six spatial principles for human-like macro placement in {\em MAGE}. 
  Each subfigure contrasts a desirable macro placement with a poor placement. 
  }
\label{fig:placement-principles}
\vspace{-12pt}
\end{figure}

\section{Principles of Human-Like Macro Placement}
\label{sec:placement-rules}

{\em MAGE} is based on the observation that expert floorplanners apply
recurring spatial principles when placing macros in an industrial PD flow. 
These principles are rarely
formalized in the literature, but are common in practice. We codify them as a
rule system that guides placement decisions in {\em MAGE}.
Figure~\ref{fig:placement-principles} illustrates the six principles.

\begin{itemize}
    \item \textit{Boundary \textcolor{black}{placement}.}
    Macros are placed along the core boundary whenever feasible, rather than
    floating in the interior. Boundary placement preserves larger contiguous
    central regions for standard-cell placement and routing, and produces more
    structured floorplans.

    \item \textit{Macro stacking.}
    Macros are organized into regular vertical or back-to-back horizontal
    pairs when dimensions and pin access permit.\footnote{\textcolor{black}{We define
    macro \textit{stacks} as a contiguous arrangement of macros 
    abutted along a shared macro edge.
    At large scale, with many macro types and different dimensions, rigid uniform-width stacking can become suboptimal; prior work~\cite{ChangCC17, Lin2019TVLSI} 
    explores more flexible arrangements. {\em MAGE} treats stacking as a preference rather than a hard rule.
    }} 
    Macros with compatible widths
    or heights may be abutted to reduce wasted whitespace and avoid fragmented
    routing channels. Spacing is introduced only when needed for pin access or
    routing resources.

    \item \textit{Pin-aware orientation.}
    Macro orientations are chosen to preserve signal-pin access. Pin edges are
    directed toward open routing space or nearby logic, while non-pin edges are
    preferentially placed against core boundaries or shared abutments. This is
    especially important in stacked placements, where adjacent macros can block
    pin access.

    \item \textit{IO keepouts.}
    Macros are placed with sufficient clearance from IO interfaces to preserve
    routing access near the chip boundary, especially along the \textit{edges} used for
    signal entry and exit.

    \item \textit{Pocket minimization.}
   A \textcolor{black}{\textit{pocket} is a \textit{free-space} region that is too
narrow or too small to be usable for standard-cell placement (formal definition in Section~\ref{sec:human_metrics}).\footnote{\textcolor{black}{Throughout this paper, we use free space to
denote the union of all regions inside the chip core 
not occupied by any macro.}}} 
    Placements should minimize pocket
area, since pockets cannot accommodate standard cells and waste
core area. \textcolor{black}{In rectilinear cores
\textcolor{black}{(cores whose boundary is an axis-aligned polygon other than a simple rectangle)}, 
\textit{dead zones} --- regions
inside the floorplan's bounding box but outside the core
polygon --- are a related but distinct concept: dead zones are
geometrically excluded from placement, whereas pockets are ``valid''
free space that is merely unusable.}

    \item \textit{Whitespace quality.}
    \textcolor{black}{Macro placements should preserve compact, contiguous, and usable whitespace for
    standard-cell placement and routing. We distinguish two conditions that {\em MAGE} tries to avoid: 
    \textit{(i)} pockets and \textit{(ii)} ``fragmentation'' of the central whitespace into multiple disconnected components. Both degrade routability and hinder downstream placement.}

\end{itemize}

Together, these principles define the structural properties that we associate
with human-like macro placement. 
\textcolor{black}{While we
evaluate {\em MAGE} on routability- and timing metrics, the
same structural regularity of human-like placements also 
benefit power delivery network
construction, clock distribution, and downstream
interpretability --- advantages that motivate human designers beyond
the metrics measured here.}
{\em MAGE} encodes them as shared rules used
throughout the placement pipeline to maintain consistent floorplanning 
intent.\footnote{The listed principles are not exhaustive; they reflect the
rules used in the current implementation. Additional expert guidance can be
added to extend the rule system to other design styles, technologies, or
constraints.}

\subsection{Human-Likeness Metrics}
\label{sec:human_likeliness_metrics}

To quantify how closely a macro placement resembles expert floorplanning,
we define four complementary metrics. Each metric is normalized to $[0,1]$,
where $1$ denotes the most human-like outcome. These metrics translate the
placement principles in Section~\ref{sec:placement-rules} into measurable
scores.\footnote{The proposed metrics are complementary rather than strictly
orthogonal. For example, the notch score measures geometric artifacts caused
by poor alignment, while the alignment score measures direct compliance with
boundary \textcolor{black}{placement}, edge alignment, corner occupation, and stacking rules.}
Let $\mathcal{M}$ denote the set of placed macros and $\mathcal{C}$ the core
region. Let $W_{\mathcal{C}}$ and $H_{\mathcal{C}}$ denote the width and
height of the bounding box of $\mathcal{C}$, respectively
(Table~\ref{tab:notation}). We use $c_w$ to denote the minimum channel width
between macros. This value is fixed before evaluation and is not tuned per
placement method.\footnote{\textcolor{black}{We set the channel width to $c_w=20\,\mu\mathrm{m}$
for NG45 and $10\,\mu\mathrm{m}$ for GF12; the finer GF12 node packs more
routing tracks per micron, so a narrower channel suffices. The IO keepout is
set to $10\,c_w$, leaving sufficient room for the standard-cell logic that
connects to the IOs.}}

\noindent\textbf{Notch score.}
A \emph{notch} is a concave step in the collective macro silhouette, often
caused by misaligned adjacent macros. We detect notches by scanning the
leftmost, rightmost, topmost, and bottommost macro extents along each axis.
A step change larger than $c_w$ between consecutive scan positions is counted
as a notch. The notch score is $S_{\mathrm{notch}} =
    \max\left(0,\,
    1 - (\sum_i d_i \ell_i) / (L_{\mathrm{total}} \cdot 2c_w)
    \right)$
where $d_i$ is the perpendicular depth of the $i$-th notch, $\ell_i$ is the
length over which the notch persists along the scanned silhouette, and
$L_{\mathrm{total}}$ is the total scanned silhouette length. For example,
when scanning the left or right silhouette along the vertical direction,
$d_i$ is measured horizontally and $\ell_i$ vertically. This metric penalizes
jagged macro silhouettes and measures compliance with the macro stacking
principle.

\noindent\textbf{Whitespace score.}
This metric measures the shape quality of free-space regions using
\textit{area-weighted rectangularity}, \textcolor{black}{where the
\textit{rectangularity} of a region is the fraction of its own
bounding box that the region fills, and the per-region values are
averaged with each region weighted by its area.} We rasterize the core at resolution $c_w/4$,
identify connected free-space regions, and compute the rectangularity of each \textcolor{black}{\textit{significant}}
region with area $A_j \geq 4c_w^2$ as
$q_j = A_j/(W_j^{\mathrm{bb}}H_j^{\mathrm{bb}})$, where
$W_j^{\mathrm{bb}}$ and $H_j^{\mathrm{bb}}$ are the width and height of the
bounding box of the $j$-th free-space region.\footnote{\textcolor{black}{Two free-space regions are
considered connected only if they share a horizontal or vertical
edge; diagonal adjacency does not count.}} 
The whitespace score is the
area-weighted mean rectangularity over all significant free-space regions: $S_{\mathrm{ws}} = (\sum_{j:\,A_j \geq 4c_w^2} q_j A_j) / (\sum_{j:\,A_j \geq 4c_w^2} A_j)$.
Higher scores indicate compact and usable whitespace, while lower scores
indicate irregular free-space regions.

\noindent\textbf{Pocket score.}
A \emph{pocket} is a small residual free-space \textit{subregion} with
limited utility for standard-cell placement or routing. We classify a 
free-space subregion as a pocket if its minimum
cross-sectional width is less than $2c_w$, or if its area is below the
minimum significant free-space threshold $4c_w^2$. The pocket score is
$S_{\mathrm{pocket}} = 1 - A_{\mathrm{pocket}}/A_{\mathrm{free}}$, where
$A_{\mathrm{pocket}}$ is the total area of all pocket subregions and
$A_{\mathrm{free}}$ is the total free area. This metric penalizes narrow or
low-utility residual spaces.

\noindent\textbf{Alignment score.}
This metric evaluates boundary \textcolor{black}{placement}, inter-macro regularity, corner
occupation, and stacking correctness: $S_{\mathrm{align}} =
    w_1 S_{\mathrm{snap}} +
    w_2 S_{\mathrm{edge}} +
    w_3 S_{\mathrm{corner}} +
    w_4 S_{\mathrm{stack}}$,
where $w_1=0.3$, $w_2=0.2$, $w_3=0.2$, and $w_4=0.3$. We use
$\tau=c_w/2$ as the geometric tolerance for all coordinate-level comparisons.
$S_{\mathrm{snap}}$ is the fraction of macros with at least one edge within
$\tau$ of the core boundary. $S_{\mathrm{edge}}$ is the fraction of macro
edges aligned with another macro edge within $\tau$. 
\textcolor{black}{$S_{\mathrm{corner}}$ is the fraction of \emph{convex}
core-boundary corners (interior angle $90^\circ$) occupied by a macro
within $\tau$; reflex corners ($270^\circ$) are excluded since they
face into the chip interior. A corner is occupied when one
macro edge lies within $\tau$ of each incident boundary segment.}

For stacking, let $\mathcal{A}$ denote the set of abutted macro pairs. Two
macros are considered abutted if their gap along one axis is less than
$\tau$ and their projections overlap along the orthogonal axis. We define
$S_{\mathrm{stack}} =
    |\{(m_i,m_j) \in \mathcal{A} :
    \mathrm{valid}(m_i,m_j)\}| / |\mathcal{A}|$
where $\mathrm{valid}(m_i,m_j)$ holds if vertically abutted macros have
matching widths within $\tau$, or horizontally abutted macros have matching
heights within $\tau$. If $\mathcal{A}$ is empty, $S_{\mathrm{stack}}$ is omitted from
$S_{\mathrm{align}}$ and the remaining alignment weights are renormalized.
This avoids rewarding placements that avoid stacking entirely.

The four metrics are combined with equal weighting to define the overall
human-likeness score: $S_{\mathrm{HM}} =
    \frac{1}{4}
    \left(
    S_{\mathrm{notch}} +
    S_{\mathrm{ws}} +
    S_{\mathrm{pocket}} +
    S_{\mathrm{align}}
    \right)$.

\noindent\textbf{Metric design and limitations.}
The proposed metrics are intended as expert-inspired structural proxies, not
as universal definitions of floorplan quality. They are designed to capture
properties that recur in expert macro \textcolor{black}{placements}. 
Some legal expert placements may receive lower
scores, for example when a rectilinear core naturally induces non-rectangular
whitespace. To reduce implementation dependence, $c_w$, the rasterization
resolution, and the alignment tolerance are fixed before evaluation and are
kept identical.

\subsection{Our Goal}
\label{sec:goal}

Given a design netlist $\mathcal{N}$, let
$\mathcal{M}=\{m_1,m_2,\ldots,m_n\}$ denote the set of hard macros with
fixed dimensions $(w_i,h_i)$, $\mathcal{S}$ the set of standard cells,
$\mathcal{C}$ the core region, and $\mathcal{I}$ the set of IO pin locations.
The core region $\mathcal{C}$ may be rectangular or rectilinear. Given an
initial macro placement
$\mathcal{X}^{0}=\{(x_i^{0},y_i^{0},o_i^{0})\}_{i=1}^{n}$, the goal is to
produce an updated placement
$\mathcal{X}=\{(x_i,y_i,o_i)\}_{i=1}^{n}$, where $(x_i,y_i)$ is the macro
location and $o_i\in\mathcal{O}$ is its orientation.
The placement must satisfy two hard \textcolor{black}{legality} constraints:
\begin{enumerate}
    \item \textit{Containment:} each macro must lie entirely inside the core
    region, i.e., $m_i \subseteq \mathcal{C}$ for all $i$.
    \item \textit{Non-overlap:} no two macros may have positive-area overlap,
i.e., $\operatorname{area}(\operatorname{int}(m_i)\cap
\operatorname{int}(m_j))=0$ for all $i\neq j$.
\end{enumerate}
Subject to these constraints, {\em MAGE} treats macro placement as a
constrained multi-objective refinement problem. The desired placement should
improve \textcolor{black}{$\mathcal{X}^{0}$ in terms of} implementation quality, 
measured by $Q_{\mathrm{PPA}}(\mathcal{X})$, and structural floorplanning quality,
measured by $S_{\mathrm{HM}}(\mathcal{X})$: 
$\mathcal{X}^{\star}
    \in
    \arg\operatorname{opt}_{\mathcal{X}}
    \left(
    Q_{\mathrm{PPA}}(\mathcal{X}),
    S_{\mathrm{HM}}(\mathcal{X})
    \right)$
subject to containment and non-overlap.
Here, $\operatorname{opt}$ denotes a procedural multi-objective refinement
rather than a fixed weighted scalarization. In the current implementation,
hard legality constraints are enforced by validation, human-like placement
principles are enforced through rule-guided generation and visual feedback,
and tournament refinement uses routed wirelength from \textcolor{black}{Cadence} 
Innovus early global
routing (eGR)~\cite{innovus} as the objective cost \textcolor{black}{($Q_{\mathrm{PPA}}$)}.
Rather than encoding the human-like placement principles as explicit penalty
terms, {\em MAGE} represents them as natural-language directives, visual
checks, and refinement criteria within the placement loop. This allows the
framework to account for both PPA objectives and structural floorplanning
quality.

\begin{figure}[htbp]
  \centering
  \includegraphics[width=\columnwidth]{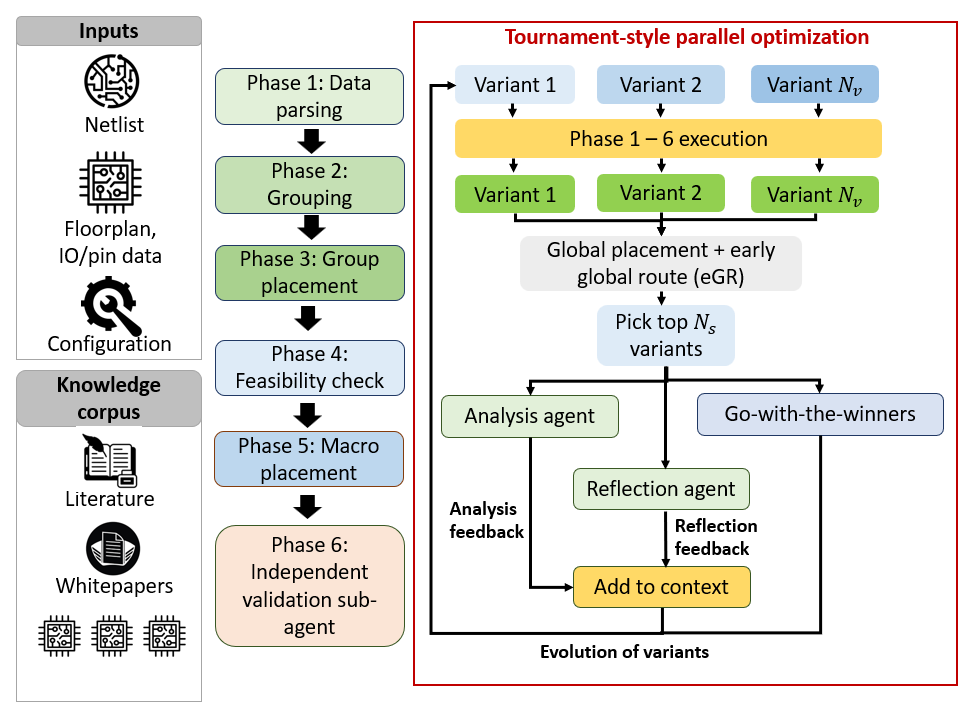}
  \caption{\textcolor{black}{Overview of {\em MAGE} framework.}}
  \label{fig:vlm_mpl_framework}
\vspace{-12pt}
\end{figure}

\section{Our Approach}
\label{sec:approach}

{\em MAGE} uses a vision-language model within a multi-agent macro placement
framework. The framework follows a ``decompose--place--validate--refine''
workflow, in which placement decisions are guided by design context and visual
feedback. This workflow is implemented through ten sub-agents (Figure~\ref{fig:vlm_mpl_framework}).

\begin{algorithm}[htbp]
\caption{Overall {\em MAGE} framework}
\label{alg:hummpl_overall}
\scriptsize
\KwInput{Netlist $\mathcal{N}$, floorplan $\mathcal{F}$, IO/pin data
$\mathcal{I}$, configuration $\Theta$, knowledge corpus $\mathcal{K}$}
\KwOutput{Final macro placement solution $\mathcal{P}^\star$}
\medskip
\tcc{\textbf{\textcolor{purple}{Phase 1: Data Parsing \& Knowledge Extraction}}}
Parse $\mathcal{N}$ and $\mathcal{F}$ to extract geometry, hierarchy,
connectivity, IO, and pin information\;
Query $\mathcal{K}$ to retrieve prior knowledge and identify known issues
in the input floorplan\;

\medskip
\tcc{\textbf{\textcolor{purple}{Phase 2: Hierarchy-Aware Grouping}}}
Construct hierarchy-aware macro groups $\mathcal{G} = \{g_1, \dots, g_k\}$
and compute inter-group connectivity\;
Identify families of similar groups
$\mathcal{Q}_1, \dots, \mathcal{Q}_p \subseteq \mathcal{G}$\;

\medskip
\tcc{\textbf{\textcolor{purple}{Phase 3: Group Placement}}}
\textcolor{black}{$\mathrm{group\_iter} \leftarrow 0$\;}
\textcolor{black}{\While{$\mathrm{group\_iter} < 5$ \textbf{and} flyline-length
reduction $\geq 2\%$}{
    Assign/refine bounding box $B_g$ for each group $g \in \mathcal{G}$
    over the core using connectivity- and IO-aware placement\;
    Render flyline visualization (Fig.~\ref{fig:visual_feedback}a);
    extract visual observations and consolidate with automated checks
    (Section~\ref{sec:vis-gates})\;
    $\mathrm{group\_iter} \leftarrow \mathrm{group\_iter} + 1$\;
}}

\medskip
\tcc{\textbf{\textcolor{purple}{Phase 4: Feasibility Checking}}}
\ForEach{group $g \in \mathcal{G}$}{
    \If{$|g| > \tau_{\max}$}{
        Recursively partition $g$ into sub-groups and re-run group placement
        within $B_g$\;
    }
    Perform feasibility checks on $g$ within $B_g$\;
    \If{feasibility fails}{
        Revise $B_g$ and repeat\;
    }
}

\medskip
\tcc{\textbf{\textcolor{purple}{Phase 5: Macro Placement}}}
$\mathcal{P} \leftarrow \emptyset$\;
\ForEach{similar-group family $\mathcal{Q}_j,\ j = 1, \dots, p$}{
    Select a reference group $g_{\mathrm{ref}} \in \mathcal{Q}_j$\;
    Generate reference placement $\mathcal{P}_{\mathrm{ref}}$ for
    $g_{\mathrm{ref}}$ guided by the principles of
    Section~\ref{sec:placement-rules}\;
    Validate $\mathcal{P}_{\mathrm{ref}}$: legality, pin access, keepouts,
    whitespace quality\;
    \If{validation fails}{
        Revise $\mathcal{P}_{\mathrm{ref}}$ or return to Phase~3\;
    }
    Transform $\mathcal{P}_{\mathrm{ref}}$ to all remaining groups in
    $\mathcal{Q}_j$ and merge into $\mathcal{P}$
    (Algorithm~\ref{alg:transform_merge})\;
}

\medskip
\tcc{\textbf{\textcolor{purple}{Phase 6: Independent Validation}}}
Perform chip-level validation on $\mathcal{P}$: overlap, boundary \textcolor{black}{placement},
IO clearance, channel spacing, boundary utilization\;
\lIf{failure due to macro placement}{return to Phase~5}
\lElseIf{failure due to group placement}{return to Phase~3}
Invoke a stateless independent auditor and execute a full audit on
$\mathcal{P}$\;
\If{audit fails}{
    Incorporate failure report and iterate up to $N_{\max}$ iterations\;
}

\medskip
\tcc{\textbf{\textcolor{purple}{Tournament-Based Solution Selection}}}
\lIf{tournament mode enabled}{$\mathcal{P}^\star \leftarrow$
\textsc{Tournament}($\mathcal{N}, \mathcal{F}, \mathcal{I},
\Theta, \mathcal{K}$) (Algorithm~\ref{alg:tournament})}
\lElse{$\mathcal{P}^\star \leftarrow \mathcal{P}$}
\Return{$\mathcal{P}^\star$}\;
\end{algorithm}

\begin{figure}[htbp]
\centering
  \includegraphics[width=1.0\columnwidth]{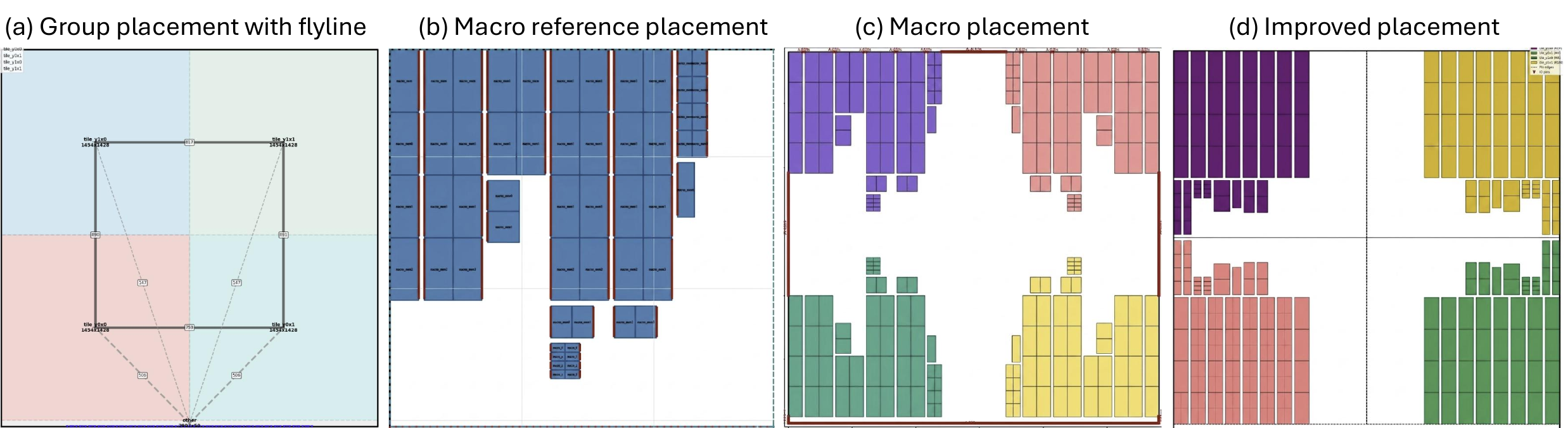}
  \caption{Progressive refinement of a macro placement through the {\em MAGE}
  pipeline on the BlackParrot (Quad-core) design. From left to right:
  \textcolor{black}{\textit{(a)}} group bounding-box assignment with flylines after Phase~3;
  \textcolor{black}{\textit{(b)}} reference macro placement for one group after Phase~5;
  \textcolor{black}{\textit{(c)}} chip-level placement after transform-and-merge across
  similar-group families; and \textcolor{black}{\textit{(d)}} final placement after
  tournament-style evolutionary optimization.}
  \label{fig:placement-flow}
\vspace{-12pt}
\end{figure}

\subsection{Framework Overview}
\label{sec:framework-overview}

\textcolor{black}{Our six-phase core} placement pipeline \textcolor{black}{applies} 
eight sub-agents:
\textit{\textbf{(1)}}~data parser,
\textit{\textbf{(2)}}~hierarchy-aware grouping,
\textit{\textbf{(3)}}~group placement,
\textit{\textbf{(4)}}~feasibility checker,
\textit{\textbf{(5)}}~macro reference placement,
\textit{\textbf{(6)}}~transform-and-merge,
\textit{\textbf{(7)}}~macro validation, and
\textit{\textbf{(8)}}~independent validation auditor.
When tournament mode is enabled (Section~\ref{sec:tournament}), two additional
sub-agents \textcolor{black}{support cross-iteration improvement:} 
\textit{\textbf{(9)}}~an analysis agent and 
\textit{\textbf{(10)}}~a reflection agent.
The input to {\em MAGE} \textcolor{black}{comprises}
\textit{(i)} an initial floorplan and macro placement from an existing tool or
manual flow,
\textit{(ii)} design netlist $\mathcal{N}$,
\textit{(iii)} IO and pin information $\mathcal{I}$,
\textit{(iv)} configuration parameters $\Theta$, and
\textit{(v)} knowledge corpus $\mathcal{K}$.
The six-phase pipeline is summarized in Alg.~\ref{alg:hummpl_overall} and
described below.

\noindent
\textbf{Lines 1--3} (Phase~1):
The data parser \textcolor{black}{\textbf{\textit{(1)}}} extracts geometry, hierarchy, connectivity, IO, and pin
information from the inputs. It also queries the knowledge corpus to retrieve
prior placement knowledge and identify floorplan issues, such as violations of
the human-like placement principles in Section~\ref{sec:placement-rules}.

\noindent
\textbf{Lines 4--6} (Phase~2):
The grouping sub-agent \textcolor{black}{\textbf{\textit{(2)}}} organizes macros into hierarchy-aware groups and
identifies families of structurally similar groups
$\mathcal{Q}_1,\ldots,\mathcal{Q}_p$. Groups within the same family later share
a reference placement, reducing repeated placement effort and promoting
consistent structure.

\noindent
\textbf{Lines 7--\textcolor{black}{12}} (Phase~3):
\textcolor{black}{Phase~3 is an iterative loop run by the group
placement sub-agent \textbf{\textit{(3)}}.} Each iteration assigns a bounding box to
every macro group over the core region using connectivity- and
IO-aware placement, with the goal of reducing inter-group
connections, preserving IO access, and maintaining usable
whitespace. \textcolor{black}{Before the iteration can complete, a
mandatory visualization gate fires (Section~\ref{sec:vis-gates}):
the sub-agent renders a flyline diagram, inspects it, and integrates
visual findings with automated checks before either accepting
the current assignment or revising it in the next iteration. The
loop terminates when the relative reduction in total weighted
flyline length falls below $2\%$ or after five iterations.}

\noindent
\textbf{Lines 13--19} (Phase~4):
The feasibility checker \textcolor{black}{\textbf{\textit{(4)}}} verifies that each group can be realized within its
assigned bounding box. It checks macro area, candidate packing patterns, and
the routing-channel spacing required for pin access. Groups exceeding a size
threshold $\tau_{\max}$ are recursively partitioned into sub-groups. If
feasibility fails, the bounding box or candidate packing pattern is revised
and the check is repeated.\footnote{We maintain several candidate packing
patterns motivated by interactions with industry experts. These patterns
include vertical stacks, horizontal rows, back-to-back pairs, L-shapes, and
U-shapes. See~\cite{mage_repo} for details.}

\noindent
\textbf{Lines 20--28} (Phase~5):
For each similar-group family $\mathcal{Q}_j$, the reference placement
sub-agent \textcolor{black}{\textbf{\textit{(5)}}} selects a reference group and generates a 
macro placement guided by
the principles in Section~\ref{sec:placement-rules}. The macro validation
sub-agent \textcolor{black}{\textbf{\textit{(7)}}} checks legality, pin access, keepouts, and whitespace quality. Once
the reference placement is accepted, the transform-and-merge sub-agent \textcolor{black}{\textbf{\textit{(6)}}} copies
its relative macro arrangement to the remaining groups in the family using
translation, reflection, or rotation as needed, and merges the resulting
placements.

\noindent
\textbf{Lines 29--35} (Phase~6):
The macro validation sub-agent \textcolor{black}{\textbf{\textit{(7)}}} checks the merged placement for overlap, boundary
\textcolor{black}{placement}, IO clearance, channel spacing, and boundary utilization. If
validation fails, control returns to Phase~5 or Phase~3 depending on the
failure type. A stateless independent auditor \textcolor{black}{\textbf{\textit{(8)}}} 
then performs a separate visual
audit. This auditor has no access to the reasoning history of the previous
phases, providing an additional check against self-consistent but flawed
solutions.\footnote{By ``self-consistent but flawed'', we mean a placement that
is internally justified by the producing phase's reasoning, but still violates
a floorplanning criterion such as pin accessibility, whitespace usability,
boundary compliance, or IO keepout.}

\noindent
\textbf{Lines 36--39}:
If tournament mode is enabled, multiple candidate placements are evaluated and
selected using Alg.~\ref{alg:tournament}. Otherwise, the current validated
placement is returned directly.

Each phase produces structured artifacts, including decision logs,
visualizations, and JSON placement files. These artifacts are passed to later
phases and provide the context needed for subsequent macro placement decisions.
The feasibility checker, reference placement agent, macro validation agent, and
independent auditor all use the placement principles \textcolor{black}{described} in
Section~\ref{sec:placement-rules}. This keeps the floorplanning rules
consistent across the pipeline. Figure~\ref{fig:placement-flow} illustrates
the progressive refinement of a macro placement through {\em MAGE}.

\subsection{Phases 1, 2: Data Parsing and Hierarchy-Aware Grouping}
\label{sec:parsing-grouping}

The data parser reads the design inputs, including macro dimensions,
floorplan \textcolor{black}{boundary}, IO locations, netlist connectivity, and master-cell pin
locations. It extracts the geometry required for downstream placement. For
rectilinear floorplans, the parser constructs the core region
$\mathcal{C}$, classifies boundary corners as convex or concave, and
identifies dead zones.
The parser also queries the knowledge corpus $\mathcal{K}$ to
retrieve relevant placement knowledge and flag floorplan issues, such as
violations of the placement principles in Section~\ref{sec:placement-rules}.
The grouping agent decomposes the design into hierarchy-aware macro groups
$\mathcal{G}=\{g_1,\ldots,g_k\}$.\footnote{\textcolor{black}{In our
experiments, $k$ typically lies between 3 to 18.}} 

\textcolor{black}{Each group is anchored by a hierarchical prefix taken from the macro
hierarchy---the path under which that group's macros reside.}
\textcolor{black}{The grouping agent then does the following. \textit{(i)} It applies the \emph{same} set
of prefixes to \emph{every} instance in the netlist, macros and
standard cells alike, assigning each to exactly one group; there is
no separate standard-cell grouping scheme.}
\textcolor{black}{\textit{(ii)} Then, the group-to-group
connection matrix is built~\cite{VidalCPGM19} by 
scanning each net in the netlist:
for a net whose pins span groups $g_a$ and $g_b$, the
$(g_a, g_b)$ entry is incremented by one; a net spanning more than
two groups increments every pairwise entry. Each entry thus counts
the number of nets shared by the two groups. Because standard cells inherit a group label from the very same
prefixes, the matrix reflects full logical connectivity rather than
macro-to-macro nets alone. E.g., a net
macro\,$\rightarrow$\,std-cell\,$\rightarrow$\,macro is recorded as a
connection between the two macros' groups, and a net that connects
\emph{only} standard cells of two groups likewise contributes to
their entry, even when no macro lies on the net. See~\cite{mage_repo} for source code
of our implementation.}
High-fanout nets with more than 100 pins are filtered to avoid distorting the
connectivity model.

The grouping agent also identifies families of \textit{similar} groups
$\mathcal{Q}_1,\ldots,\mathcal{Q}_p \subseteq \mathcal{G}$. Two groups are
considered similar if they have the same macro count and matching macro types
and dimensions. Groups in the same family later share a reference macro
placement, as described in Section~\ref{sec:macro-placement}. Finally, IO pins
are classified by core edge, and their connections to macro groups are tracked
for IO-proximity-aware group placement in Phase~3.

\subsection{Phase 3: Group Placement}
\label{sec:group-placement}

The group placement agent assigns \textcolor{black}{\textit{group bounding boxes}}, i.e., bounding box $B_g$ for each macro
group $g \in \mathcal{G}$ \textcolor{black}{within} the core region
$\mathcal{C}$. The objective is to reduce \textit{weighted flyline
length} \textcolor{black}{subject to two constraints: \textit{(i)} preserving
IO access, and \textit{(ii)} tiling the entire core, i.e., the group bounding
boxes partition $\mathcal{C}$ with no gaps and no overlaps
($\bigcup_{g \in \mathcal{G}} B_g = \mathcal{C}$, with pairwise
disjoint interiors).}\footnote{We define weighted flyline length as
$\sum_{(g_i,g_j)} w_{ij}\,\|c_i-c_j\|_1$, where $w_{ij}$ is the
connection weight between groups $g_i$ and $g_j$ from the
group-to-group connection matrix, and $c_i,c_j$ are the centroids of
their assigned bounding boxes. IO-to-group flylines are handled
similarly using IO-cluster centroids.} For rectilinear floorplans,
tiling is performed over the true core polygon rather than its
bounding box. We refer to each boundary segment of $\mathcal{C}$ as
a \textit{core edge}.

Before group placement, IO pins are clustered by proximity. Groups are ranked by
IO affinity, and groups with strong IO connectivity are assigned to
IO-adjacent regions first. For core edges with dense IO clusters, {\em MAGE}
reserves wider routing channels between the IO pins and nearby group boxes.
\textcolor{black}{Then,} inter-group placement is guided by weighted flylines between group centroids,
with weights derived from the group-to-group connection matrix from Phase~2.
Strongly connected groups are placed near each other.

Phase~3 \textcolor{black}{itself} proceeds iteratively. In each iteration, the agent: \textit{(i)} proposes a complete
group-box assignment, \textit{(ii)} runs validation scripts, \textit{(iii)} renders a visualization
and extracts observations from the rendered image, and \textit{(iv)} produces a structured report
before revising or accepting the placement. The visualization includes core
boundaries, group boxes, IO pins, group-to-group and IO-to-group flylines, and
a coverage overlay that highlights any unclaimed core area
(Figure~\ref{fig:placement-flow}). The process terminates when
\textit{(i)} all validation checks pass, \textit{(ii)} the relative reduction
in weighted flyline length falls below 2\% between consecutive iterations, or
\textit{(iii)} the maximum iteration count is reached.\footnote{
In our experiments, the maximum
number of group-placement iterations is set to five.}

\subsection{Phase 4: Feasibility Check}
\label{sec:feasibility}

Before macro coordinates are generated for a group, the feasibility checker
verifies that the group can be physically realized within its assigned
bounding box $B_g$. This is a mandatory gate: Phase~5 proceeds only after all
groups pass feasibility. Groups with $|g|>\tau_{\max}$ are recursively
partitioned into \textit{sub-groups} by re-invoking Phases~2--4 within the
parent group's bounding box. Groups with $10<|g|\leq\tau_{\max}$ may be
partitioned if needed, while groups with $|g|\leq10$ proceed without
partitioning.\footnote{In our implementation, $\tau_{\max}=30$.} The resulting
sub-group bounding boxes tile the parent group's bounding box area.

For each group or sub-group $g$, let $B_g$ have width $W_g$ and height $H_g$.
Under a candidate packing pattern, suppose the pattern requires $n_x$ packing
columns with effective widths $\{\hat{w}_1,\ldots,\hat{w}_{n_x}\}$ and $n_y$
packing rows with effective heights $\{\hat{h}_1,\ldots,\hat{h}_{n_y}\}$. A
packing column (resp. row) is a vertical (resp. horizontal) track containing
one or more macros, and its effective width (resp. height) is the maximum macro
width (resp. height) among macros in that track. Let $c_x$ and $c_y$ denote
the horizontal and vertical channel spacings, and let \textcolor{black}{$\delta_{\mathrm{io},x}$ and
$\delta_{\mathrm{io},y}$ denote the total IO keepout margin consumed
along the horizontal and vertical axes of $B_g$, respectively.
Concretely, $\delta_{\mathrm{io},x} = d_{\mathrm{io}}^{L} +
d_{\mathrm{io}}^{R}$ is the sum of the left- and right-side keepouts,
and $\delta_{\mathrm{io},y} = d_{\mathrm{io}}^{B} +
d_{\mathrm{io}}^{T}$ is the sum of the bottom- and top-side keepouts,
where each side's keepout $d_{\mathrm{io}}^{\,\cdot}$ is the
mandatory clearance from any IOs on that side.} Feasibility requires
\begin{equation}
\begin{aligned}
    \textstyle\sum_{i=1}^{n_x} \hat{w}_i + (n_x-1)c_x + \delta_{\mathrm{io},x}
    &\leq W_g, \\
    \textstyle\sum_{j=1}^{n_y} \hat{h}_j + (n_y-1)c_y + \delta_{\mathrm{io},y}
    &\leq H_g.
\end{aligned}
\label{eq:feasibility}
\end{equation}
For sides adjacent to dense IO clusters, the reserved
IO margin is doubled. If either inequality is violated, the checker reduces
channel spacing, changes the packing pattern, or requests a revised bounding
box from Phase~3.

\begin{algorithm}[t]
\caption{Transform-and-merge procedure}
\label{alg:transform_merge}
\scriptsize
\KwInput{Validated reference \textcolor{black}{macro} placement
$\mathcal{P}_{\mathrm{ref}}$\textcolor{black}{; similar-group
family $\mathcal{Q}_j$ with designated reference group
$g_{\mathrm{ref}} \in \mathcal{Q}_j$}; group bounding boxes
$\{B_g\}_{g \in \mathcal{Q}_j}$; IO pin data $\mathcal{I}$}
\KwOutput{Merged chip-level \textcolor{black}{macro} placement $\mathcal{P}$}
\medskip
\tcc{\textbf{\textcolor{purple}{Reference Initialization}}}
$\mathcal{P} \leftarrow \mathcal{P}_{\mathrm{ref}}$\;
\medskip
\tcc{\textbf{\textcolor{purple}{Per-Group Transformation}}}
\ForEach{target group $g \in \mathcal{Q}_j\setminus\{g_{\mathrm{ref}}\}$}{
    Determine spatial relation between $B_{g_{\mathrm{ref}}}$ and $B_g$\;
    Select transformation $T_g \in \{$translation, mirror-X, mirror-Y,
    rotation-$180^\circ\}$\;
    \ForEach{macro $m \in \mathcal{P}_{\mathrm{ref}}$}{
        Identify the corresponding macro
$m'$ in target group $g$ using the matched macro type and index within the
similar-group family \textcolor{black}{$\mathcal{Q}_j$}\;
        Apply $T_g$ to the coordinates of $m$ and assign the transformed coordinates
to $m'$\;
Remap the orientation of $m'$ under $T_g$\;
    }
    Verify containment: all transformed macros lie within $B_g$\;
    Verify boundary \textcolor{black}{placement} and stacking relations are preserved\;
    Verify pin accessibility and IO keepout compliance w.r.t.\ $\mathcal{I}$\;
    \If{any constraint is violated}{
        \Return{FAIL and return to Phase~5}\;
    }
    Let $\mathcal{P}'_g$ denote the transformed placement for $g$\;
    $\mathcal{P} \leftarrow \mathcal{P} \cup \mathcal{P}'_g$\;
}
\medskip
\tcc{\textbf{\textcolor{purple}{Chip-Level Validation}}}
Check $\mathcal{P}$ for cross-group overlaps, insufficient channel
spacing, and boundary violations\;
Perform boundary-utilization audit: edge coverage, corner occupation,
and central whitespace quality\;
\lIf{any chip-level check fails}{\Return{FAIL}}
\Return{$\mathcal{P}$}\;
\end{algorithm}

\subsection{Phase 5: Macro Placement}
\label{sec:macro-placement}

Phase~5 consists of three steps: reference placement, transform-and-merge, and validation.

\noindent\textbf{Reference placement.}
For each similar-group family $\mathcal{Q}_j$, the reference placement
agent selects one reference group $g_{\mathrm{ref}} \in \mathcal{Q}_j$
and generates a macro placement $\mathcal{P}_{\mathrm{ref}}$.
\textcolor{black}{This $(g_{\mathrm{ref}}, \mathcal{P}_{\mathrm{ref}})$
pair is the input that Alg.~\ref{alg:transform_merge}
(transform-and-merge) consumes when propagating
$\mathcal{P}_{\mathrm{ref}}$ to the remaining groups in
$\mathcal{Q}_j$.} This placement serves as the template for the
remaining groups in the family \textcolor{black}{(see~\cite{mage_repo} for
details)}.

The placement follows a boundary-first strategy consistent with
Section~\ref{sec:placement-rules}. Candidate boundary slots are generated from
core corners, core edges, and group edges. Slots are ranked by their ability to
preserve central whitespace, maintain IO and pin access, and support regular
stacking. Larger macros are assigned to higher-ranked slots first. Macro
orientations are chosen using signal-pin locations. For each macro, edges that
contain signal pins are directed toward routing channels or nearby \textcolor{black}{whitespace}, while edges with no signal pins are preferentially placed against core
boundaries or shared abutments. Arrangement patterns, such as vertical stacks,
\textcolor{black}{same-row abutments}, L-shapes, U-shapes, and grids, are selected based on group
geometry, boundary position, IO proximity, macro mix, and connectivity.\footnote{In
our implementation, the prompt includes a library of example arrangement
patterns, but the agent may compose new arrangements when they better match the
group geometry and connectivity.} The resulting placement is written in a JSON
format used by the visualization and validation scripts.

\noindent\textbf{Transform-and-merge.}
After $\mathcal{P}_{\mathrm{ref}}$ passes validation, the transform-and-merge
step maps it to the remaining groups in $\mathcal{Q}_j$;
\textcolor{black}{see Alg.~\ref{alg:transform_merge}}. The chip-level placement
$\mathcal{P}$ is initialized with the validated reference placement
(Lines~1--2). For each target group $g$, the spatial relationship between
$B_{g_{\mathrm{ref}}}$ and $B_g$ determines the transformation, e.g.,
translation, reflection, or $180^\circ$ rotation (Lines~5--6). Macro coordinates and
orientations are remapped accordingly (Lines~7--10). The transformed placement
is then checked for containment, non-overlap, boundary placement, pin
accessibility, IO keepout compliance, and whitespace quality (Lines~11--13).
If any check fails, control returns to the reference placement step
(Lines~14--15).
After all groups are merged, cross-group non-overlap,
boundary compliance, channel spacing, and whitespace quality are checked (Line~19). A
boundary-utilization audit further checks core-edge coverage, corner
occupation, and fragmentation of the remaining whitespace (Line~20).

\noindent\textbf{Validation.}
The validation step uses both automated checks and visual review. Automated
scripts check legality, pin access, IO keepouts, whitespace quality, and
post-merge correctness with respect to the human-likeness principles in
Section~\ref{sec:placement-rules}, producing PASS/FAIL results with violation
details. The rendered placement is then reviewed to detect spatial issues that
may not be captured by numerical checks. Any FAIL item triggers a feedback
loop: macro arrangement failures return to reference placement, while group
bounding-box failures return to Phase~3. This ensures that local
macro placement errors and higher-level group-placement errors are corrected at
the appropriate stage\textcolor{black}{~\cite{mage_repo}}.

\begin{figure}[htbp]
\centering
  \includegraphics[width=0.9\columnwidth]{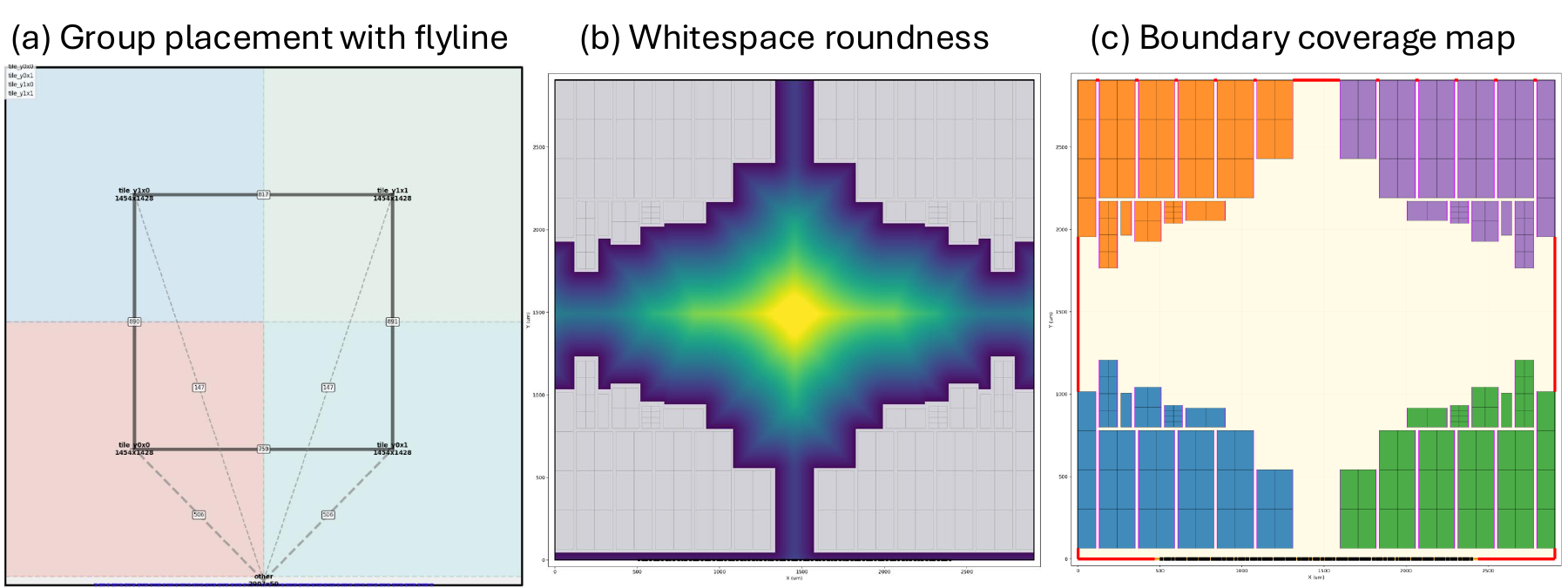}
  \caption{Some visual artifacts used for feedback in {\em MAGE}.
  \textit{(a)}~Group-placement flyline visualization.
  \textit{(b)}~Whitespace roundness heatmap; purple indicates fragmented 
  whitespace with low roundness.
  \textit{(c)}~Boundary-coverage map; red lines mark boundary gaps. Design: \textcolor{black}{BlackParrot}.}
  \label{fig:visual_feedback}
\vspace{-12pt}
\end{figure}

\subsection{Phase 6: Independent Validation Sub-Agent}
\label{sec:independent-validation}

Phase~6 performs an independent audit of 
the post-merge macro placement
$\mathcal{P}$. Although Phase~5 already checks post-merge correctness, that
validation is part of the same generation loop that produced the placement.
Phase~6 therefore invokes a separate validation agent \textcolor{black}{that has} no access to the
reasoning history of Phases~1--5. This provides a second-pass audit from a
fresh context, using only the rendered artifacts and validation reports. The
goal is to catch residual issues that may survive the integrated validation
loop, such as blocked pin access, insufficient IO keepout, poor boundary
compliance, narrow routing channels, or fragmented whitespace.

\textcolor{black}{The Phase~5 validator
and the Phase~6 auditor differ in three
respects: the Phase~5 validator inherits the full reasoning history
of Phases~1--5 and shares the generators' goal of producing a passing
macro placement, while the Phase~6 auditor is \emph{stateless} (sees only
$\mathcal{P}$, validation script outputs, and rendered images) and is
prompted to find what the in-loop validator missed.}
The auditor runs the validation scripts from Phase~5 and reviews the generated
artifacts: \textit{(i)} PASS/FAIL reports, \textit{(ii)} placement
visualizations with macro pin edges and highlighted violations, \textit{(iii)}
whitespace diagnostics, and \textit{(iv)} textual feedback
(Figure~\ref{fig:visual_feedback}). It then produces a visual feedback report
with specific rearrangement suggestions for affected groups or sub-groups
(Figure~\ref{fig:visual_feedback_report}).
A regression guard tracks placement quality across validation iterations to
detect cases where fixing one issue degrades another. If violations remain
after $N_{\max}$ iterations, the unresolved failures are documented with
root-cause analysis, and the best valid or least-violating iteration is
selected as the final output. Feedback from the independent auditor returns to
Phase~5 for macro arrangement issues and to Phase~3 for group placement issues.

\begin{algorithm}[t]
\caption{Tournament-style optimization}
\label{alg:tournament}
\footnotesize
\KwInput{Design inputs $(\mathcal{N}, \mathcal{F}, \mathcal{I})$,
configuration $\Theta$, knowledge corpus $\mathcal{K}$, number of
variants $N_v$, survivor count $N_s$, number of tournament rounds $R$}
\KwOutput{Best macro placement solution \textcolor{black}{found,} $\mathcal{P}^\star$}
\medskip
\tcc{\textbf{\textcolor{purple}{Initialization}}}
Initialize variant pool $\mathcal{V}^{(1)}=\{v_1,\ldots,v_{N_v}\}$ with
distinct seeds, contexts, and improvement suggestions\;
Initialize evaluation history $\mathcal{H}\leftarrow\emptyset$\;

\medskip
\tcc{\textbf{\textcolor{purple}{Tournament Loop}}}
\For{round $r \leftarrow 1$ \KwTo $R$}{
    \tcc{\textcolor{purple}{Parallel placement and evaluation}}
    \ForEach{variant $v \in \mathcal{V}^{(r)}$}{
        Run Phases~1--6 of {\em MAGE} with tournament mode disabled to obtain
    \textcolor{black}{macro} placement $\mathcal{P}_v$\;
    Run Innovus \textcolor{black}{global placement and} early global routing on $\mathcal{P}_v$\;
    Record eGR routed wirelength $\mu_v$\;
    $\mathcal{H}\leftarrow\mathcal{H}\cup\{(\mathcal{P}_v,\mu_v)\}$\;
    }
    \tcc{\textcolor{purple}{Selection}}
    Rank all variants by $\mu_v$\;
    Select the top-$N_s$ survivors
    $\mathcal{W}^{(r)} \subset \mathcal{V}^{(r)}$\;
    \tcc{\textcolor{purple}{Analysis and reflection (sub-agents~9--10)}}
    \ForEach{survivor $w \in \mathcal{W}^{(r)}$}{
        Invoke analysis agent (sub-agent~9): compare $\mathcal{P}_w$
        against a reference placement, extract placement
        discrepancies $\{d_1, d_2, \ldots\}$ and compute
        per-discrepancy HPWL attribution
        \textcolor{black}{$\{\Delta_1, \Delta_2, \ldots\}$ such that
        $\sum_k \Delta_k = \mathrm{HPWL}(\mathcal{P}_w) -
        \mathrm{HPWL}(\mathcal{P}_{\mathrm{ref}})$}\;
        Invoke reflection agent (sub-agent~10): evaluate whether prior
        suggestions helped or failed, generate revised guidance for
        round $r+1$\;
    }
    \tcc{\textcolor{purple}{Go-with-the-winners replication}}
    Build $\mathcal{V}^{(r+1)}$ by replicating survivors proportionally
to rank until $|\mathcal{V}^{(r+1)}|=N_v$\;
    Attach parent context, analysis feedback, and reflection feedback to
    each replicated variant \textcolor{black}{in $\mathcal{V}^{(r+1)}$}\;
}

\medskip
\tcc{\textbf{\textcolor{purple}{Solution Selection}}}
$\mathcal{P}^\star \leftarrow
\arg\min_{(\mathcal{P},\mu)\in\mathcal{H}} \mu$\;
\Return{$\mathcal{P}^\star$}\;
\end{algorithm}

\subsection{Tournament-Style Parallel Optimization}
\label{sec:tournament}

{\em MAGE} supports a tournament-style optimization mode inspired by
Go-With-The-Winners~\cite{AldousV94}. Alg.~\ref{alg:tournament}
summarizes the procedure. A pool of $N_v$ placement variants is initialized,
and an evaluation history $\mathcal{H}$ stores all evaluated placements and
their eGR routed wirelength values (Lines~1--3). In each round, all variants
independently run the full {\em MAGE} pipeline in parallel. Each resulting
macro placement is evaluated by running Innovus eGR and measuring the rWL ($\mu_v$) from
eGR (Lines~4--11). 
\textcolor{black}{We consider
$\mu_v$ only because eGR
is the latest stage in the flow that runs relatively fast per variant
(full post-route signoff would inflate tournament cost by an order
of magnitude per round) and because eGR rWL is empirically a strong
proxy for downstream PPA metrics (Table~\ref{tab:results_ppa}).}
The variants are ranked by the rWL from eGR, and the
top-$N_s$ variants are selected as survivors (Lines~12--14).
For each survivor, an analysis step compares its placement against a 
\textcolor{black}{single} reference
placement, namely the input commercial macro placer solution. The analysis estimates how much each placement difference contributes to HPWL
by recomputing HPWL after isolating the affected groups or macros. It considers
differences such as \textit{(i)} distances between connected groups,
\textit{(ii)} spatial spread of macros within a group, and \textit{(iii)}
nearest-neighbor relationships among macros (Lines~15--17). 
A reflection step then identifies which prior suggestions were
useful and generates revised guidance for the next round (Line~18). The
next-round pool is constructed by GWTW replication: higher-ranked survivors
are replicated more often until the pool size reaches $N_v$.\footnote{In our
implementation, {\em MAGE} maintains six variants i.e., $N_v=6$. The best
variant is copied three times, the second-best twice, and the third-best
once.} Each replicated
variant inherits its parent context, analysis feedback, and reflection feedback
(Lines~19--21). Finally, the placement with the lowest eGR rWL in $\mathcal{H}$ is returned as $\mathcal{P}^\star$ (Lines~22--24).

\section{Multimodal Reasoning and Visual Feedback}
\label{sec:vision-use}

A design principle of {\em MAGE} is that placement quality should be assessed
through both automated checks and visual feedback. Automated scripts enforce
hard constraints such as overlap, boundary, and channel-spacing violations.
However, several properties of expert floorplans are spatial, including
stacking regularity, pin accessibility, whitespace organization, and overall
visual coherence. {\em MAGE} therefore uses visual inspection as a mandatory
part of the placement framework.
\textcolor{black}{Vision is used in two ways as described below, with some representative 
visual artifacts shown in Figure~\ref{fig:visual_feedback}.} 


\subsection{Mandatory Visualization Gates}
\label{sec:vis-gates}

Every iteration in Phase~3 and Phase~5 follows a mandatory visual gating
sequence. Before proceeding, the corresponding sub-agent must: (i) generate a
placement visualization, (ii) run automated checks, (iii) inspect the rendered image,
(iv) extract visual observations, and (v) produce a structured report.\footnote{\textcolor{black}{The rendering scripts,
the visual-only-fact extraction prompts, and the visual quality
checks are available in~\cite{mage_repo}.}}
In Phase~3, the visualization is a group flyline diagram showing group
bounding boxes, IO pins, and weighted inter-group flylines
\textcolor{black}{(Figure~\ref{fig:visual_feedback}(a))}. 
This image is used to assess whether
strongly connected groups are placed near each other and whether unnecessary
flyline crossings remain.
In Phase~5, the visualization shows individual macros, group membership, macro
pin edges, group bounding boxes, the core boundary, and highlighted
violations. It is paired with whitespace diagnostics, including a roundness
heatmap, a distance-to-center map, and a boundary-coverage map \textcolor{black}{(see Figures~\ref{fig:visual_feedback}(b) and (c))}. 
These views help identify violated human principles.

Each visual review must include \emph{visual-only facts} i.e., observations derived
strictly from the rendered layout rather than from numerical audit data alone. This
requirement ensures that the image is explicitly inspected rather than only
summarized from script outputs. The review also evaluates three visual quality
criteria: whitespace shape quality, pin visibility, and floorplan regularity.
A visual failure triggers a return to the appropriate placement phase even if
all automated checks pass.
When validation returns to an earlier phase, {\em MAGE} generates side-by-side
comparisons of the current and previous placements. The report must document
visual improvements, visual regressions, and any accepted tradeoffs. A
regression guard tracks iteration-to-iteration metrics to prevent a fix in one
criterion from degrading another.
\begin{figure}[t]
\centering
\begin{tcolorbox}[colback=gray!5, colframe=gray!60, fonttitle=\small\bfseries, title={\em MAGE} Visual Feedback Report (sub-agent~8{,} excerpt), fontupper=\scriptsize, left=4pt, right=4pt, top=2pt, bottom=2pt, boxrule=0.5pt, arc=2pt]

\textbf{Placement audit observations:}

\hangindent=1em\hangafter=1
$\bullet$ The placement uses a 4-tile quadrant layout. Macros are
concentrated along the top, left, and right edges, with a large open
area in the center and bottom.

\hangindent=1em\hangafter=1
$\bullet$ Pin-edge lines are visible on most macros and generally face
toward open channels or the chip interior. No cases of pin edges facing
directly into adjacent macros.


\medskip
\textbf{Whitespace diagnostic observations:}

\hangindent=1em\hangafter=1
$\bullet$ \textit{Roundness map:} The free space forms a cross-shaped
pattern centered on the chip. Roundness is 0.086 ($< 0.3$) because the
whitespace has narrow corridors extending between macro stacks rather
than a compact central region.


\hangindent=1em\hangafter=1
$\bullet$ \textit{Boundary coverage:} Top edge $\sim$76\% covered
(good), left/right edges $\sim$66\% each (moderate), bottom edge 0\%
(correct---all 135 IOs are on this edge).

\medskip
\textbf{Actionable rearrangement suggestion:}

\hangindent=1em\hangafter=1
$\bullet$ \textit{Fix IO keepout violations (158 failures):} Shift all
macros in the bottom two tiles upward by $\geq$10$\mu$m (from
lly=50$\mu$m to lly$\geq$60$\mu$m). This resolves all 158 violations
without changing the relative macro arrangement.

\end{tcolorbox}
\caption{Excerpt of a visual feedback report generated by the independent
validation auditor. The report converts rendered placement images and
whitespace diagnostics into image-derived observations and actionable
refinement suggestions for the next placement iteration. Design: \textcolor{black}{BlackParrot.}}
\label{fig:visual_feedback_report}
\end{figure}

\subsection{Auxiliary Vision Pipeline}
\label{sec:aux-vision}
\textcolor{black}{The visualization gates of
Section~\ref{sec:vis-gates} ensure each placement iteration is
inspected before it proceeds. The auxiliary vision pipeline serves a
complementary purpose: rather than gating a single iteration, it
converts selected visual artifacts into structured natural-language
feedback that is written into the shared context and consumed by
downstream sub-agents and later iterations, letting visual
observations propagate without re-passing raw images at every
prompt.}

\textcolor{black}{The pipeline reuses the artifacts already rendered
by the visualization gates (e.g., Phase~3 flyline diagrams, Phase~5
macro, etc.) but emits
text rather than images. Each artifact is summarized into
observations and actionable suggestions. During tournament
optimization, for instance, it flags macros lying far from the
centroids of their respective sets of connected standard cells and
tracks how these displacements evolve across iterations.}

When the auxiliary VLM detects a quality issue, it produces a
structured feedback document --- the visual observation, the affected
groups or macros, and a suggested refinement
(Figure~\ref{fig:visual_feedback_report}). \textcolor{black}{The
feedback is routed to the phase responsible for the issue}: Phase~3
for group-level issues, Phase~5 for macro-arrangement issues, \textcolor{black}{or} the
tournament analysis step for cross-variant improvement. Rendered
artifacts are thus converted into structured guidance for subsequent
refinement, rather than used only for final inspection.

\begin{figure}[htbp]
\centering
  \includegraphics[width=0.8\columnwidth]{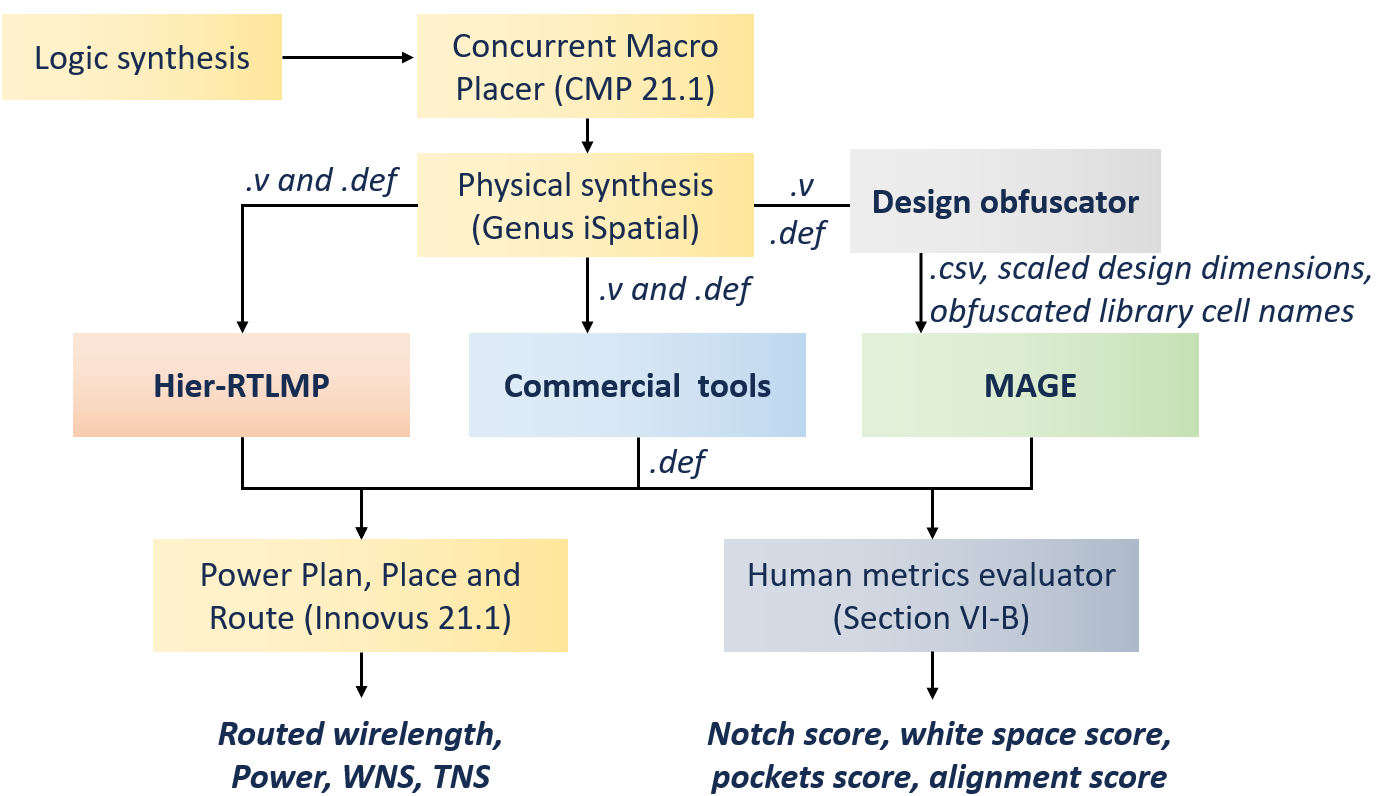}
  \caption{Evaluation flow for macro placements produced by different macro placers.
  \textcolor{black}{The \emph{Design obfuscator} block prevents tool-specific or
enablement-specific information from being exposed to the LLM.}}
  \label{fig:eval_flow}
\end{figure}


\section{Experimental Setup and Results}
\label{sec:experiments-results}

We implement {\em MAGE} in Python and use Anthropic Claude Opus
4.6~\cite{claude} as the VLM backbone for all agents. The auxiliary vision
pipeline (Section~\ref{sec:aux-vision}) uses a separate VLM instance for image
analysis. All experiments are run on a server with eight 2.4\,GHz Intel
Xeon(R) Gold 6148 processors and 376\,GB RAM.
We evaluate {\em MAGE} on nine benchmark designs across two technology
enablements: \textit{(i)} Ariane~\cite{ariane}, BlackParrot
(Quad-core)~\cite{blackparrot}, and MemPool Group~\cite{mempool} in
NanGate45 \textcolor{black}{(NG45)}~\cite{ng45,TILOS-repo}; and \textit{(ii)} Ariane, BlackParrot,
Tabla01, Tabla05, Tabla09, and Tabla13~\cite{EsmaeilzadehGGGKK21} in
GlobalFoundries 12nm (GF12). To evaluate generalization, we further include
two unseen\footnote{Unseen designs are not used during prompt development,
knowledge-corpus construction, or manual tuning of the {\em MAGE}
configuration $\Theta$. They are evaluated only after the framework is
finalized.} GF12 designs, VTA~\cite{BanerjeeBCDJK21} and
GeneSys01~\cite{EsmaeilzadehGGGKK21}.
We compare {\em MAGE} against three baselines:
\textit{(i)} Comm-1 and \textit{(ii)} Comm-2, \textcolor{black}{two releases of an unnamed}
commercial macro placer, and \textit{(iii)} OpenROAD's
Hier-RTLMP~\cite{Kahng2024HierRTLMP,hier-rtlmp}.\footnote{The commercial
baselines are included to contextualize solution quality, not as product
benchmarks. \textcolor{black}{Hier-RTLMP does not complete some of the benchmarks.}}

\noindent\textbf{Evaluation flow.}
Figure~\ref{fig:eval_flow} shows the end-to-end evaluation methodology, which
follows~\cite{ChengKKWW23, ChengKKWW25}. First, each RTL design is synthesized
with Cadence Genus~21.1 to produce a gate-level netlist. The synthesized
netlist is then loaded into Cadence Innovus~21.1~\cite{innovus}, 
where the commercial macro
placer generates an initial macro placement. The resulting floorplan DEF is
passed to the Cadence Genus iSpatial flow for physical-aware synthesis and
initial standard-cell placement.
Macro placement solutions are then generated using Comm-1, Comm-2,
Hier-RTLMP, and {\em MAGE}, where available.\footnote{We do not pass DEF, LEF,
Verilog, or other design files directly to {\em MAGE}. Instead, a reference
parser extracts only the required macro, standard-cell, floorplan, IO, and
connectivity information into CSV files. All dimensions are scaled, and macro
master names are anonymized before being provided to the VLM. This
obfuscation step, shown in Figure~\ref{fig:eval_flow}, prevents tool-specific
or enablement-specific information from being exposed.} For each macro
placement solution, the floorplan DEF is loaded into Innovus~21.1 for
place-and-route. Standard cells are reset to unplaced, macro locations are
legalized using \texttt{refine\_macro\_placement}, and the flow proceeds
through power delivery network generation, placement, clock tree synthesis,
routing, and post-route optimization.
From each post-routed design, we extract routed wirelength (rWL), total power,
worst negative slack (WNS), and total negative slack (TNS). 
\textcolor{black}{Note that this
post-route rWL is the final reported wirelength and is distinct from
the eGR wirelength used as the in-loop ranking
metric $\mu_v$ during tournament optimization
(Section~\ref{sec:tournament}).}
In parallel, each
macro placement is evaluated by a human-likeness metric evaluator
(Section~\ref{sec:human_likeliness_metrics}), which reports overall human-likeness scores.

We organize the results into four parts:
\textit{(i)} PPA evaluation (Section~\ref{sec:eval_ppa});
\textit{(ii)} human-likeness evaluation (Section~\ref{sec:human_metrics});
\textit{(iii)} ablation studies on the knowledge corpus $\mathcal{K}$ and
visual feedback (Section~\ref{sec:ablation}); and
\textit{(iv)} additional case studies on anonymized designs, unseen designs, rectilinear floorplans, and utilization sweep (Section~\ref{sec:case_studies}).

\begin{table}[t]
\caption{Post-route PPA comparison across NG45 and GF12 designs. Below
each design name, the triplet (Inst., Macros, Types) reports the total
instance count, the number of hard macros, and the number of unique macro
types, respectively. NG45 values are absolute (rWL in mm, Pwr in mW, WNS
in ps, TNS in ns); GF12 values are normalized as described in
Section~\ref{sec:eval_ppa}. Human and Hier-RTLMP baselines are available
only for NG45. The best value per design is shown in \textbf{bold}.}
\label{tab:results_ppa}
\centering
\scriptsize
\setlength{\tabcolsep}{3pt}
\renewcommand{\arraystretch}{0.95}
\begin{tabular}{@{}lllrrrr@{}}
\toprule
\textbf{Tech.} & \textbf{Design} & \textbf{Method} &
\textbf{rWL} & \textbf{Pwr} & \textbf{WNS} & \textbf{TNS} \\
\midrule

\multirow{15}{*}{NG45}
& \multirow{5}{*}{\shortstack[l]{Ariane\\\scriptsize(109K, 133, 1)}}
 & Human       & 4681 & 832 & -88 & -46.8 \\
& & Comm-1      & 4057 & 852 & -154 & -196.5 \\
& & Comm-2      & \textbf{3994} & \textbf{830} & -144 & -127.9 \\
& & Hier-RTLMP  & 5297 & 836 & -165 & -223.4 \\
& & {\em MAGE}  & 4424 & 832 & \textbf{-81} & \textbf{-32.7} \\
\cmidrule(lr){2-7}

& \multirow{5}{*}{\shortstack[l]{BP\\\scriptsize(736K, 220, 6)}}
 & Human       & 25916 & 4470 & -97 & -321.9 \\
& & Comm-1      & \textbf{23144} & 4429 & -144 & -356.2 \\
& & Comm-2      & 24080 & 4447 & -182 & -842.0 \\
& & Hier-RTLMP  & 28866 & 4512 & -150 & -456.6 \\
& & {\em MAGE}  & 26987 & 4479 & \textbf{-88} & \textbf{-139.7} \\
\cmidrule(lr){2-7}

& \multirow{5}{*}{\shortstack[l]{MemPool\\\scriptsize(2.57M, 324, 4)}}
 & Human       & 107598 & 2640 & -49 & -11.9 \\
& & Comm-1      & \textbf{102907} & \textbf{2587} & -20 & -1.0 \\
& & Comm-2      & 104674 & \textbf{2587} & \textbf{-14} & -8.7 \\
& & Hier-RTLMP  & 114073 & 2695 & -62 & -4.9 \\
& & {\em MAGE}  & 107309 & 2636 & -32 & \textbf{-0.8} \\

\midrule

\multirow{18}{*}{GF12}
& \multirow{3}{*}{\shortstack[l]{Tabla01\\\scriptsize(320K, 760, 7)}}
 & Comm-1      & 1.000 & \textbf{1.000} & -0.28 & -1080.0 \\
& & Comm-2      & \textbf{0.985} & 1.002 & -0.25 & -769.3 \\
& & {\em MAGE}  & 1.391 & 1.027 & \textbf{-0.22} & \textbf{-463.3} \\
\cmidrule(lr){2-7}

& \multirow{3}{*}{\shortstack[l]{Tabla05\\\scriptsize(550K, 2488, 7)}}
 & Comm-1      & 1.000 & 1.000 & -0.20 & -276.0 \\
& & Comm-2      & \textbf{0.985} & \textbf{0.991} & -0.18 & -73.8 \\
& & {\em MAGE}  & 1.445 & 1.086 & \textbf{-0.15} & \textbf{-15.8} \\
\cmidrule(lr){2-7}

& \multirow{3}{*}{\shortstack[l]{Tabla09\\\scriptsize(185K, 368, 8)}}
 & Comm-1      & 1.000 & 1.000 & -0.19 & -98.5 \\
& & Comm-2      & \textbf{0.994} & \textbf{0.985} & -0.18 & -182.5 \\
& & {\em MAGE}  & 1.458 & 1.071 & \textbf{-0.15} & \textbf{-48.5} \\
\cmidrule(lr){2-7}

& \multirow{3}{*}{\shortstack[l]{Tabla13\\\scriptsize(352K, 1232, 8)}}
 & Comm-1      & 1.000 & 1.000 & -0.19 & -216.9 \\
& & Comm-2      & \textbf{0.995} & \textbf{0.999} & \textbf{-0.15} & -125.8 \\
& & {\em MAGE}  & 1.589 & 1.081 & -0.18 & \textbf{-72.9} \\
\cmidrule(lr){2-7}

& \multirow{3}{*}{\shortstack[l]{Ariane\\\scriptsize(93K, 133, 1)}}
 & Comm-1      & \textbf{1.000} & 1.000 & -0.16 & -142.3 \\
& & Comm-2      & 1.031 & \textbf{0.993} & -0.12 & -106.4 \\
& & {\em MAGE}  & 1.223 & 1.021 & \textbf{-0.10} & \textbf{-68.0} \\
\cmidrule(lr){2-7}

& \multirow{3}{*}{\shortstack[l]{BP\\\scriptsize(665K, 196, 6)}}
 & Comm-1      & 1.000 & 1.000 & -0.09 & -138.9 \\
& & Comm-2      & \textbf{0.987} & \textbf{1.000} & -0.14 & -225.5 \\
& & {\em MAGE}  & 1.099 & 1.010 & \textbf{-0.08} & \textbf{-33.4} \\
\bottomrule
\end{tabular}
\end{table}

\begin{table}[t]
\caption{Geometric-mean improvement of {\em MAGE} over each baseline.
Positive values indicate improvement by {\em MAGE}. rWL and Pwr are
reported as percentage reduction; WNS and TNS are reported as percentage
improvement.}
\label{tab:geomean}
\centering
\scriptsize
\setlength{\tabcolsep}{4pt}
\renewcommand{\arraystretch}{1.1}
\begin{tabular}{@{}l rrrr@{}}
\toprule
\textbf{{\em MAGE} vs.} & \textbf{rWL (\%)} & \textbf{Pwr (\%)}
  & \textbf{WNS (\%)} & \textbf{TNS (\%)} \\
\midrule
Human        & \textbf{0.6}   & \textbf{0.1}  & \textbf{18.3} & \textbf{72.5} \\
Comm-1       & -29.7 & -3.8 & \textbf{19.3} & \textbf{70.0} \\
Comm-2       & -29.7 & -4.5 & \textbf{11.1} & \textbf{74.0} \\
Hier-RTLMP   & \textbf{9.8}   & \textbf{1.1}  & \textbf{47.0} & \textbf{80.4} \\
\bottomrule
\end{tabular}
\vspace{-10pt}
\end{table}

\begin{figure}[htbp]
\centering
\scriptsize

\textbf{(a) NG45} \\[5pt]
\resizebox{0.8\columnwidth}{!}{%
\begin{tabular}{@{}c @{\hspace{1pt}} c@{\hspace{1pt}}c@{\hspace{1pt}}c@{\hspace{1pt}}c@{\hspace{1pt}}c@{}}
& Human & Comm-1 & Comm-2 & Hier-RTLMP & {\em MAGE} \\[3pt]
\rotatebox{90}{\makebox[1.4cm]{\small Ariane}} &
\includegraphics[height=1.4cm]{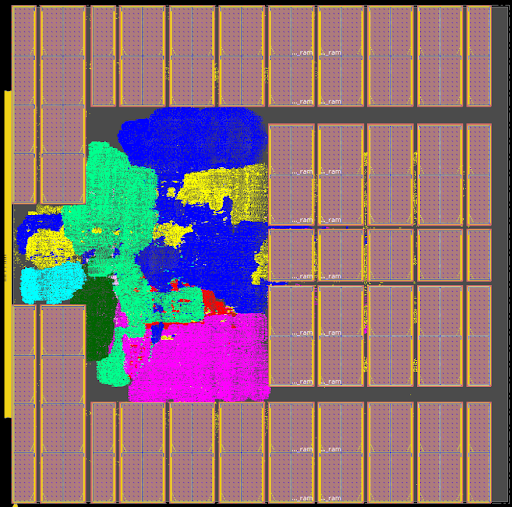} &
\includegraphics[height=1.4cm]{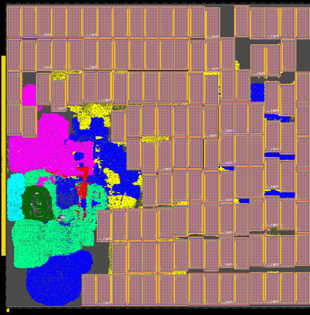} &
\includegraphics[height=1.4cm]{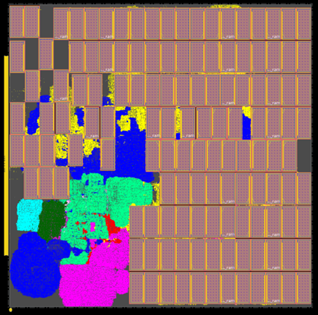} &
\includegraphics[height=1.4cm]{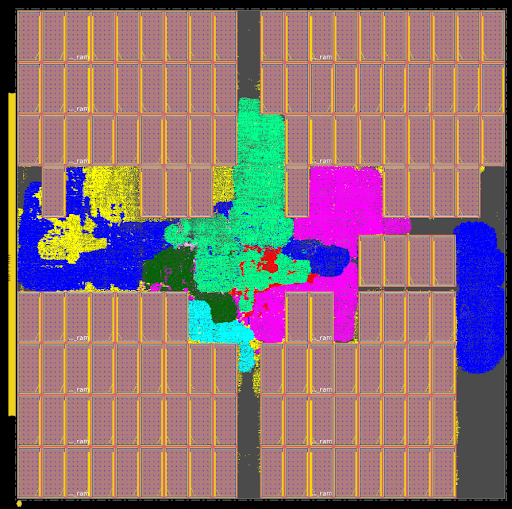} &
\includegraphics[height=1.4cm]{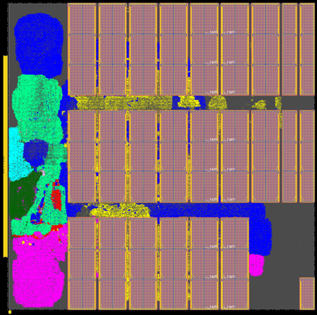} \\[2pt]
\rotatebox{90}{\makebox[1.4cm]{\small BP}} &
\includegraphics[height=1.4cm]{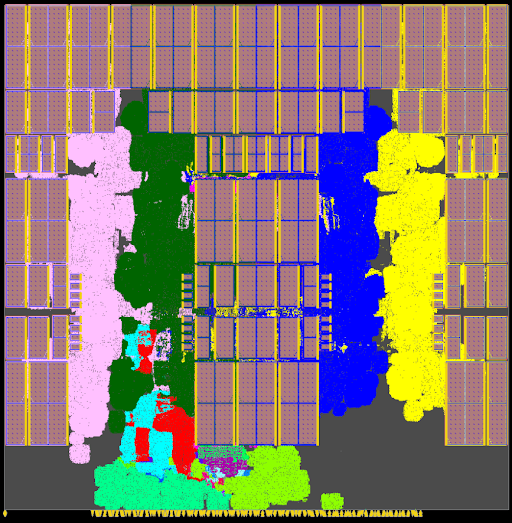} &
\includegraphics[height=1.4cm]{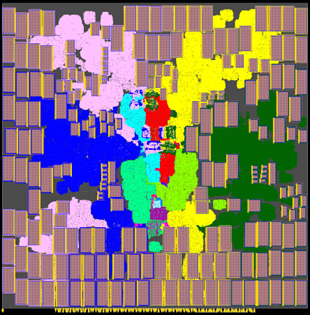} &
\includegraphics[height=1.4cm]{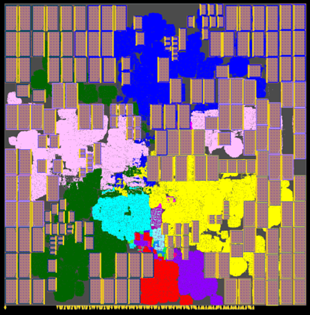} &
\includegraphics[height=1.4cm]{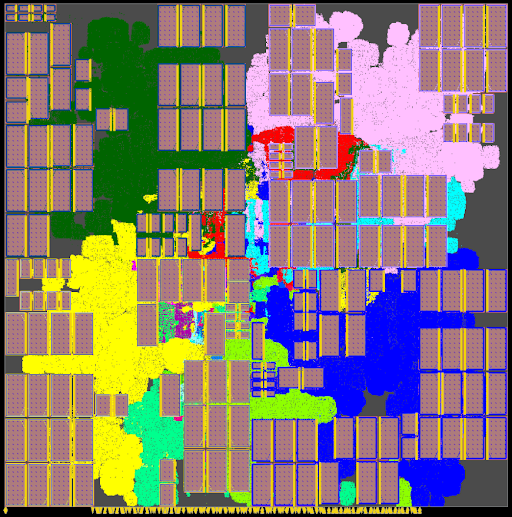} &
\includegraphics[height=1.4cm]{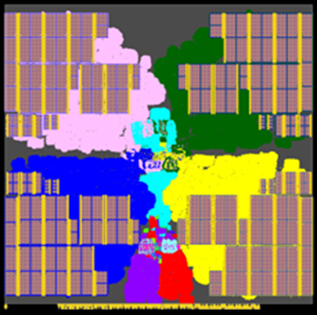} \\[2pt]
\rotatebox{90}{\makebox[1.4cm]{\small MemPool}} &
\includegraphics[height=1.4cm]{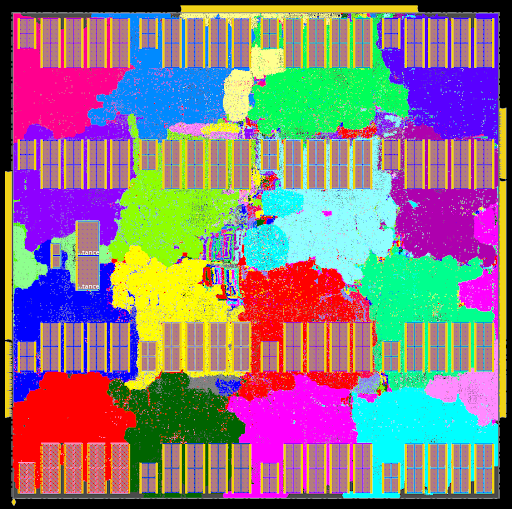} &
\includegraphics[height=1.4cm]{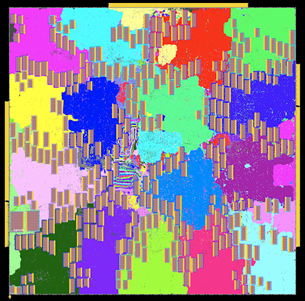} &
\includegraphics[height=1.4cm]{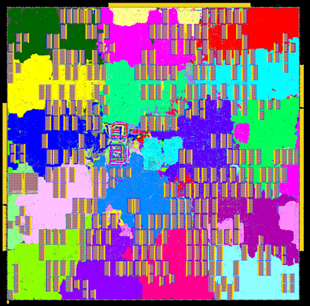} &
\includegraphics[height=1.4cm]{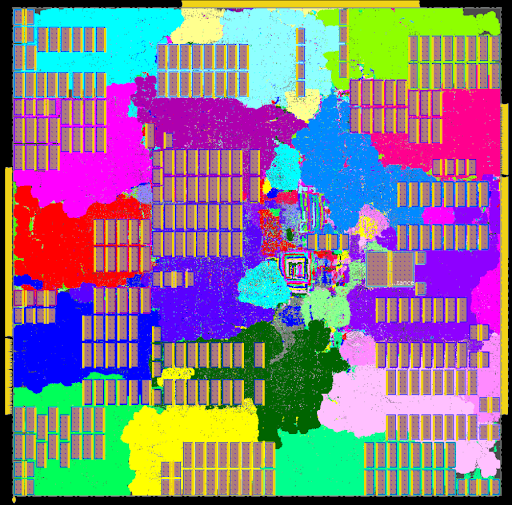} &
\includegraphics[height=1.4cm]{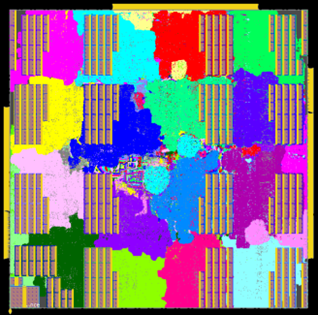} \\
\end{tabular}%
}


\textbf{(b) GF12} \\[5pt]
\resizebox{0.6\columnwidth}{!}{%
\begin{tabular}{@{}c @{\hspace{2pt}} c@{\hspace{2pt}}c@{\hspace{2pt}}c@{}}
& Comm-1 & Comm-2 & {\em MAGE} \\[3pt]
\rotatebox{90}{\makebox[1.8cm]{\footnotesize Tabla01}} &
\includegraphics[height=2.0cm]{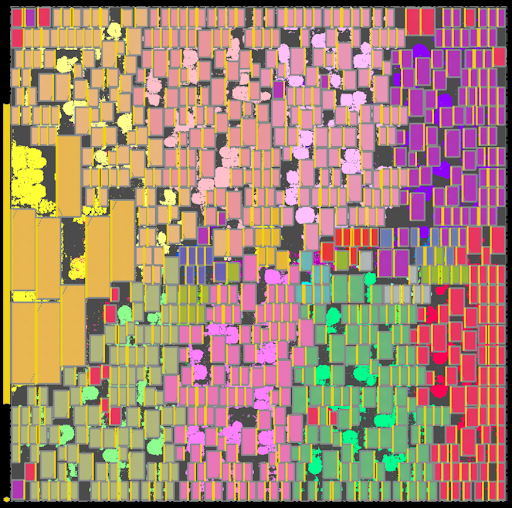} &
\includegraphics[height=2.0cm]{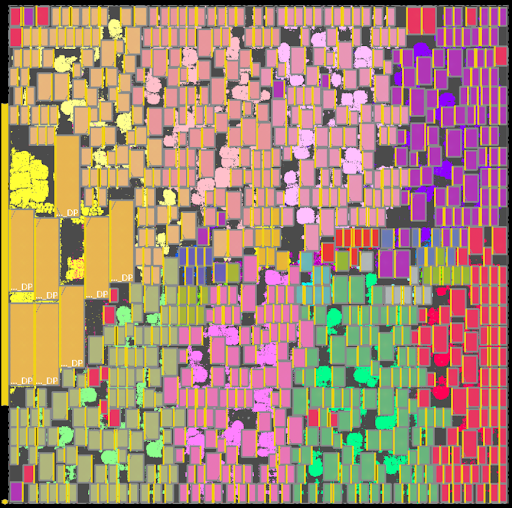} &
\includegraphics[height=2.0cm]{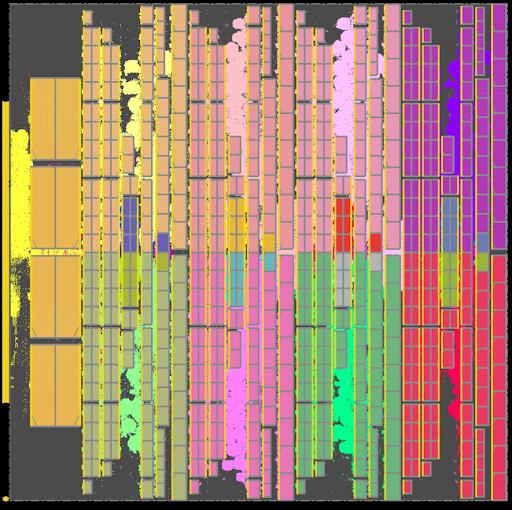} \\[2pt]
\rotatebox{90}{\makebox[1.8cm]{\footnotesize Tabla05}} &
\includegraphics[height=2.0cm]{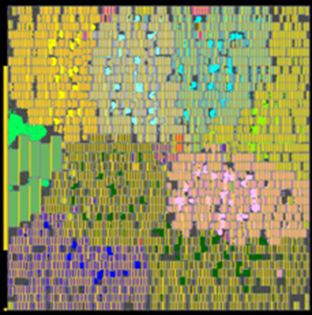} &
\includegraphics[height=2.0cm]{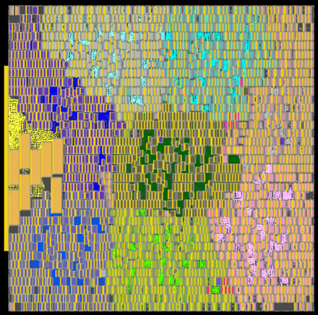} &
\includegraphics[height=2.0cm]{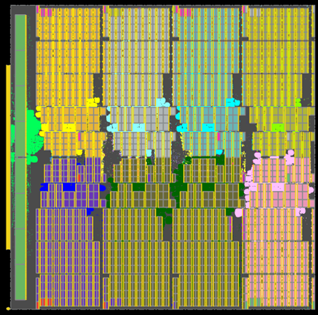} \\[2pt]
\rotatebox{90}{\makebox[1.8cm]{\footnotesize Tabla09}} &
\includegraphics[height=2.0cm]{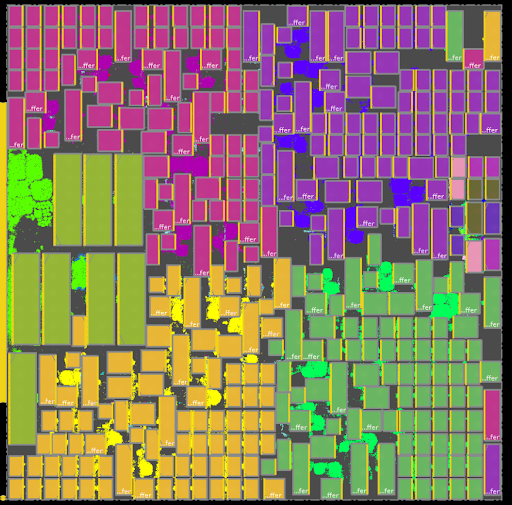} &
\includegraphics[height=2.0cm]{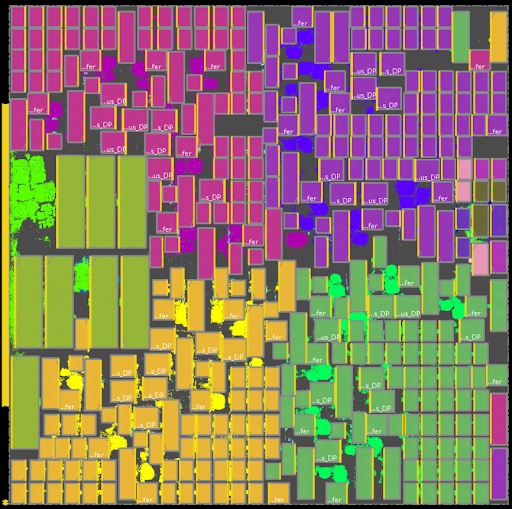} &
\includegraphics[height=2.0cm]{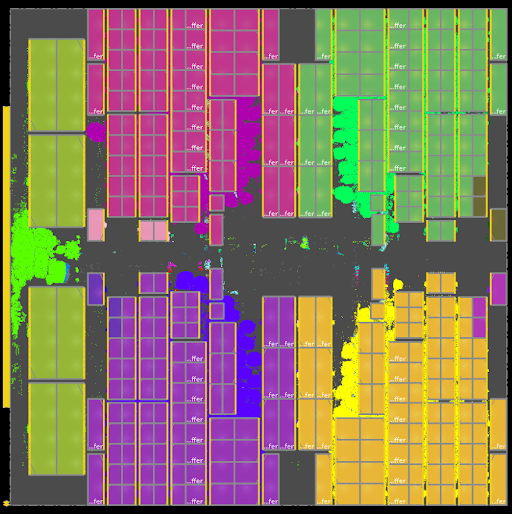} \\[2pt]
\rotatebox{90}{\makebox[1.8cm]{\footnotesize Tabla13}} &
\includegraphics[height=2.0cm]{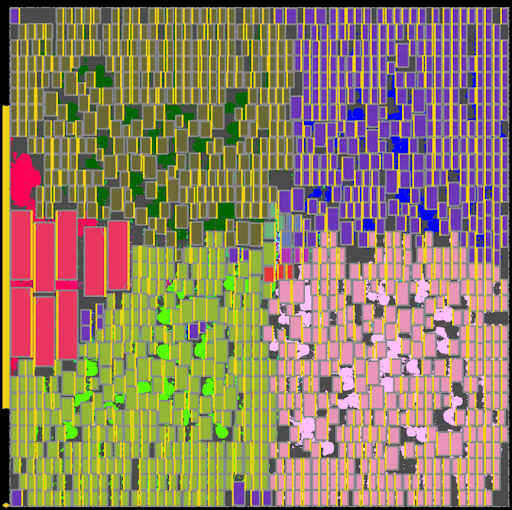} &
\includegraphics[height=2.0cm]{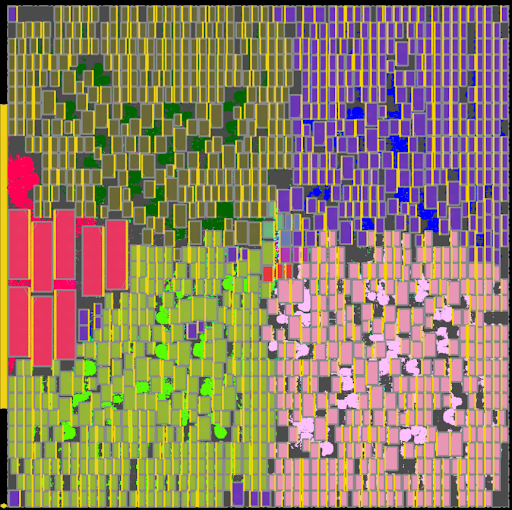} &
\includegraphics[height=2.0cm]{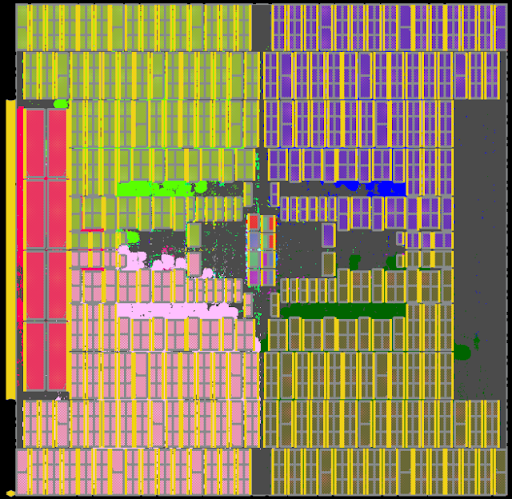} \\[2pt]
\rotatebox{90}{\makebox[1.8cm]{\footnotesize Ariane}} &
\includegraphics[height=2.0cm]{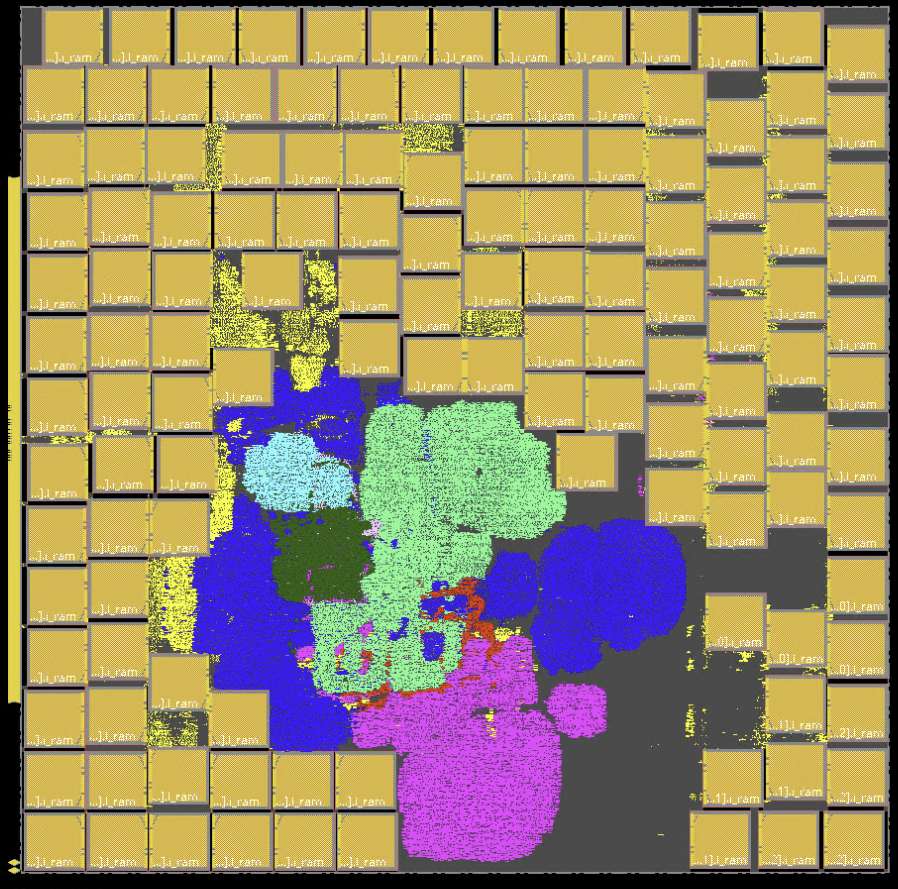} &
\includegraphics[height=2.0cm]{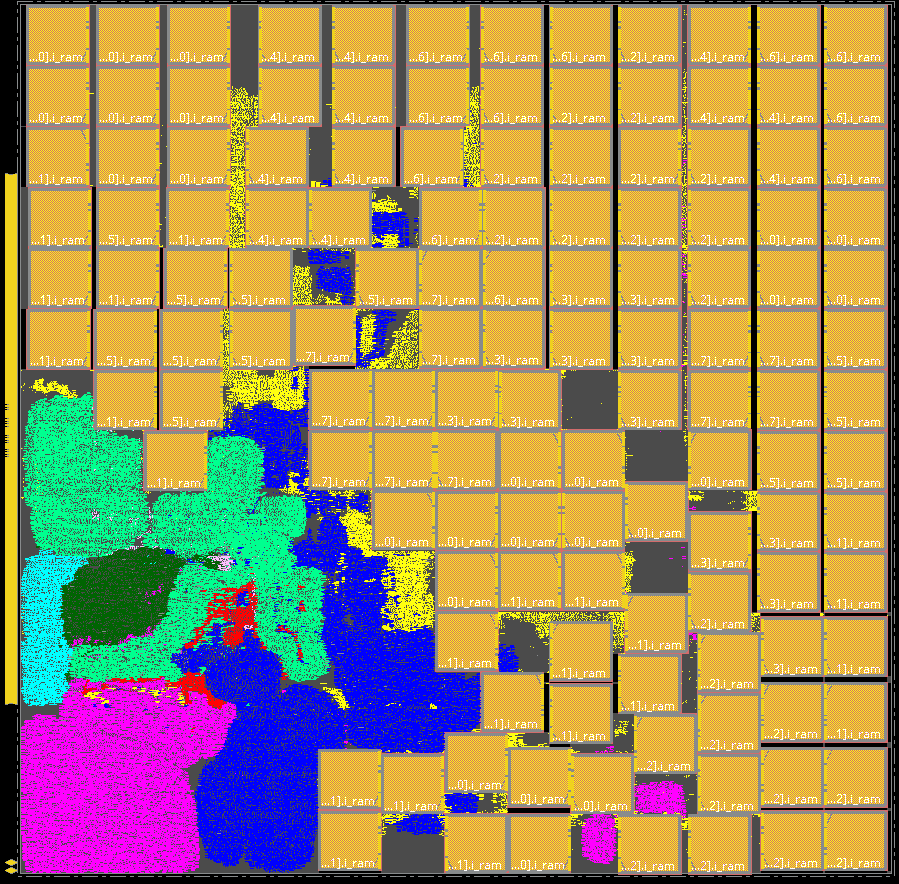} &
\includegraphics[height=2.0cm]{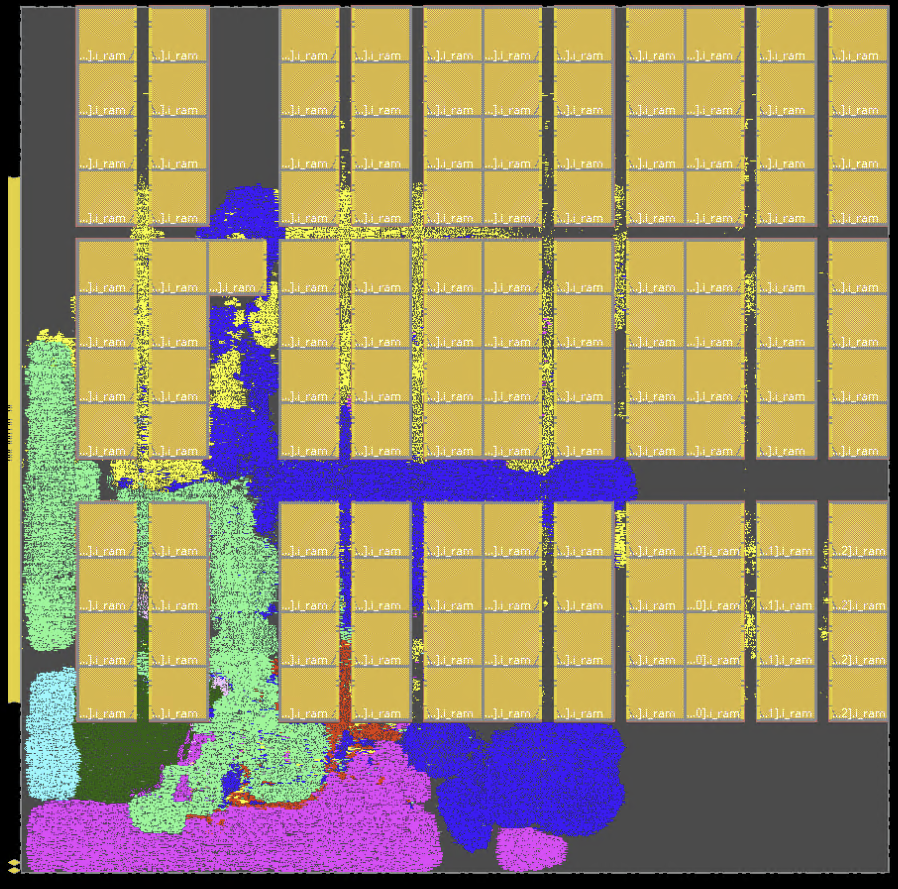} \\[2pt]
\rotatebox{90}{\makebox[1.8cm]{\footnotesize BP}} &
\includegraphics[height=2.0cm]{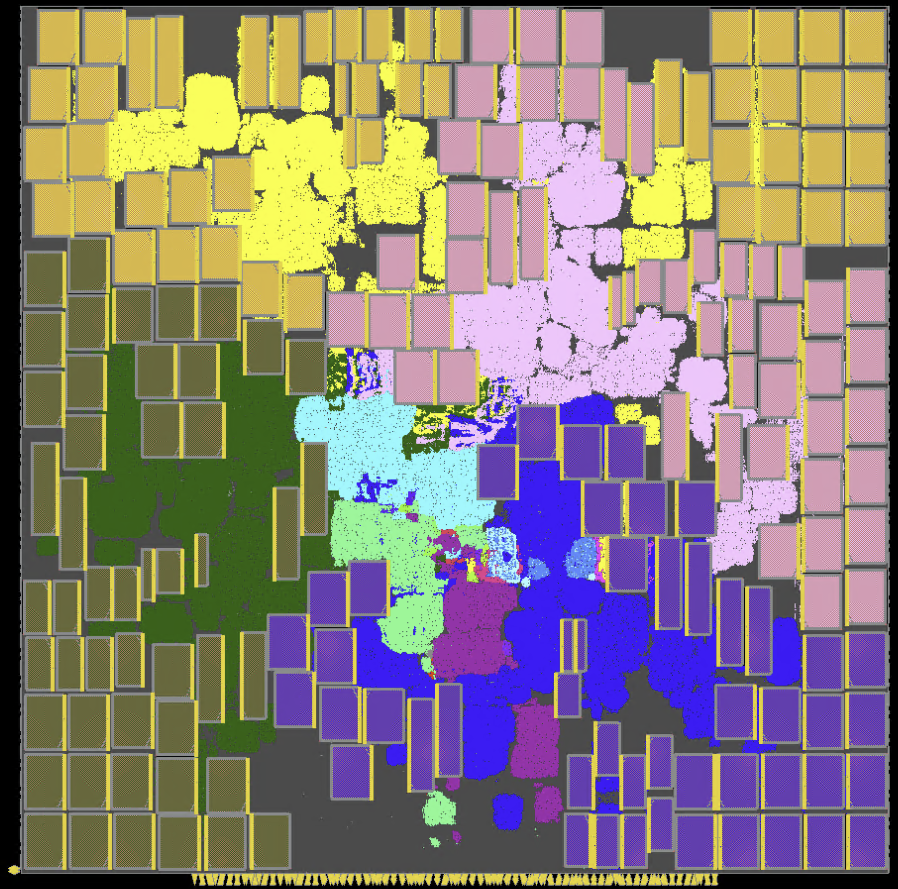} &
\includegraphics[height=2.0cm]{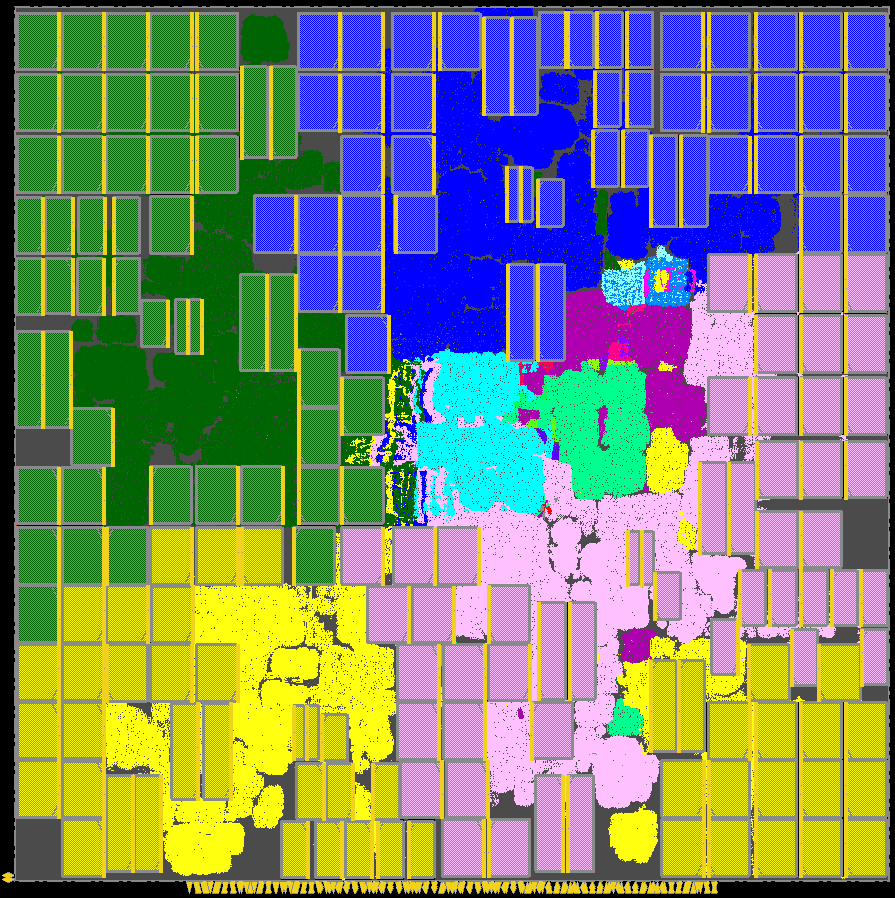} &
\includegraphics[height=2.0cm]{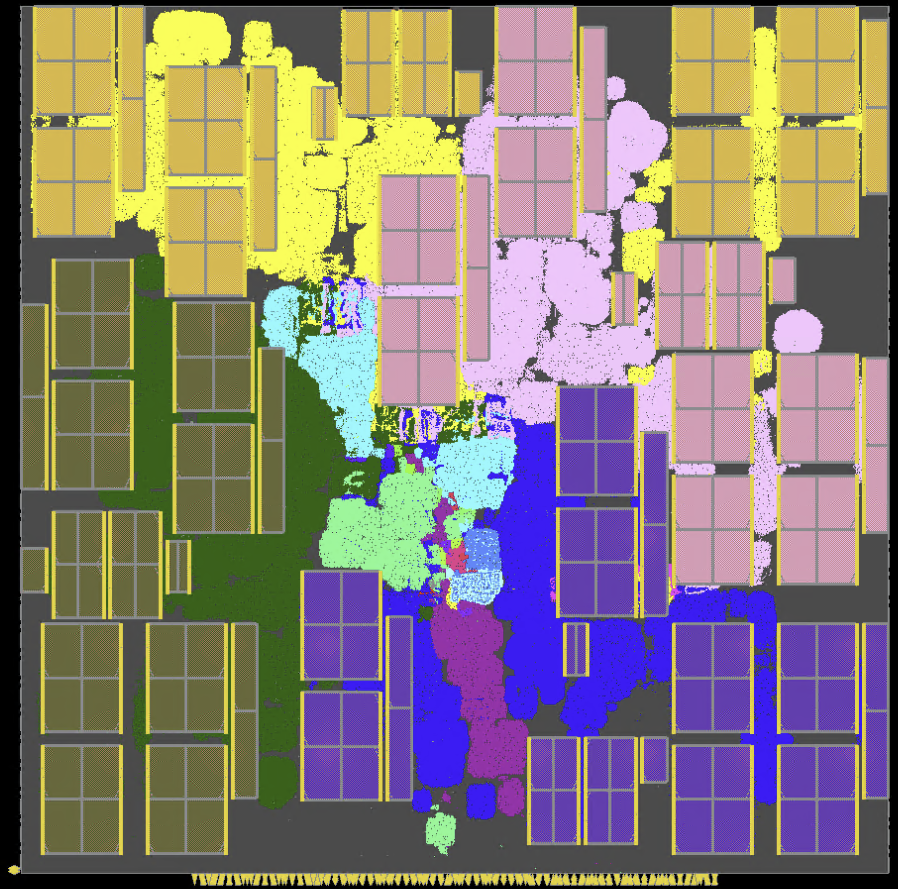} \\
\end{tabular}%
}

\caption{Macro placement comparisons across NG45 and GF12 enablements.
(a) NG45 designs are compared across Human, Comm-1, Comm-2, Hier-RTLMP,
and {\em MAGE}. (b) GF12 designs are compared across Comm-1, Comm-2,
and {\em MAGE}.}
\label{fig:placement-grid}
\end{figure}

\subsection{Evaluation of PPA}
\label{sec:eval_ppa}

Table~\ref{tab:results_ppa} reports post-route PPA for the NG45 and GF12
enablements. Because GF12 is a proprietary enablement, all GF12 numbers
\textcolor{black}{that we report} are normalized: rWL and Pwr are normalized to the
Comm-1 result of each design, and WNS and TNS are normalized to the target
clock \textcolor{black}{period.} NG45 numbers are reported as absolute values.
Table~\ref{tab:geomean} summarizes geometric-mean improvements,
and Figure~\ref{fig:placement-grid} shows the corresponding macro placements.
Across the evaluated baselines, {\em MAGE} improves WNS by 11.1\%--47.0\% and
TNS by 70.0\%--80.4\% in geometric mean. The main improvement is in timing:
\textit{(i)} relative to the human expert baseline, {\em MAGE} achieves similar rWL and power with better WNS and TNS, 
\textit{(ii)} relative to Hier-RTLMP, it improves all reported
PPA metrics, and 
\textit{(iii)} relative to the commercial baselines, it trades higher rWL and
power for better timing, especially TNS.

\noindent
\textbf{{\em MAGE} vs.\ Human:}
On the NG45 designs, {\em MAGE} improves geometric-mean WNS by 18.3\% and TNS
by 72.5\% over the human expert baseline, while maintaining comparable rWL and
power. The improvement is consistent across the designs. The largest TNS gain
occurs on MemPool, where TNS improves by 93\%. These results indicate that
{\em MAGE} can \textcolor{black}{achieve} key physical characteristics of expert floorplans while
improving post-route timing.

\noindent
\textbf{{\em MAGE} vs.\ Comm-1:}
Against Comm-1, {\em MAGE} improves geometric-mean WNS by 19.3\% and TNS by
70.0\%, at the cost of 29.7\% higher rWL and 3.8\% higher power. The TNS
improvement is observed across all evaluated designs. The tradeoff is more
\textcolor{black}{visible in} GF12, where {\em MAGE} incurs higher rWL but reduces 
TNS by 57\%--94\% \textcolor{black}{on Tabla designs.} 

\noindent
\textbf{{\em MAGE} vs.\ Comm-2:}
Compared with Comm-2, {\em MAGE} improves geometric-mean WNS by 11.1\% and
TNS by 74.0\%, with 29.7\% higher rWL and 4.5\% higher power. {\em MAGE}
improves TNS on all designs, and WNS also
improves in most cases, with two exceptions. On MemPool, Comm-2 achieves the
best WNS ($-14$\,ps), while {\em MAGE} achieves the best TNS
($-0.8$\,ns). On Tabla13, Comm-2 achieves better WNS, while {\em MAGE}
achieves better TNS. 

\noindent
\textbf{{\em MAGE} vs.\ Hier-RTLMP:}
The largest gains are observed against Hier-RTLMP on NG45. {\em MAGE}
improves geometric-mean WNS by 47.0\% and TNS by 80.4\%, while also reducing
rWL by 9.8\% and power by 1.1\%. Unlike the \textcolor{black}{comparisons with Comm-1 
and Comm-2}, this is not a timing--wirelength tradeoff: {\em MAGE} improves every
reported PPA metric over Hier-RTLMP.

\begin{figure}[htpb]
\centering
  \includegraphics[width=0.6\columnwidth]{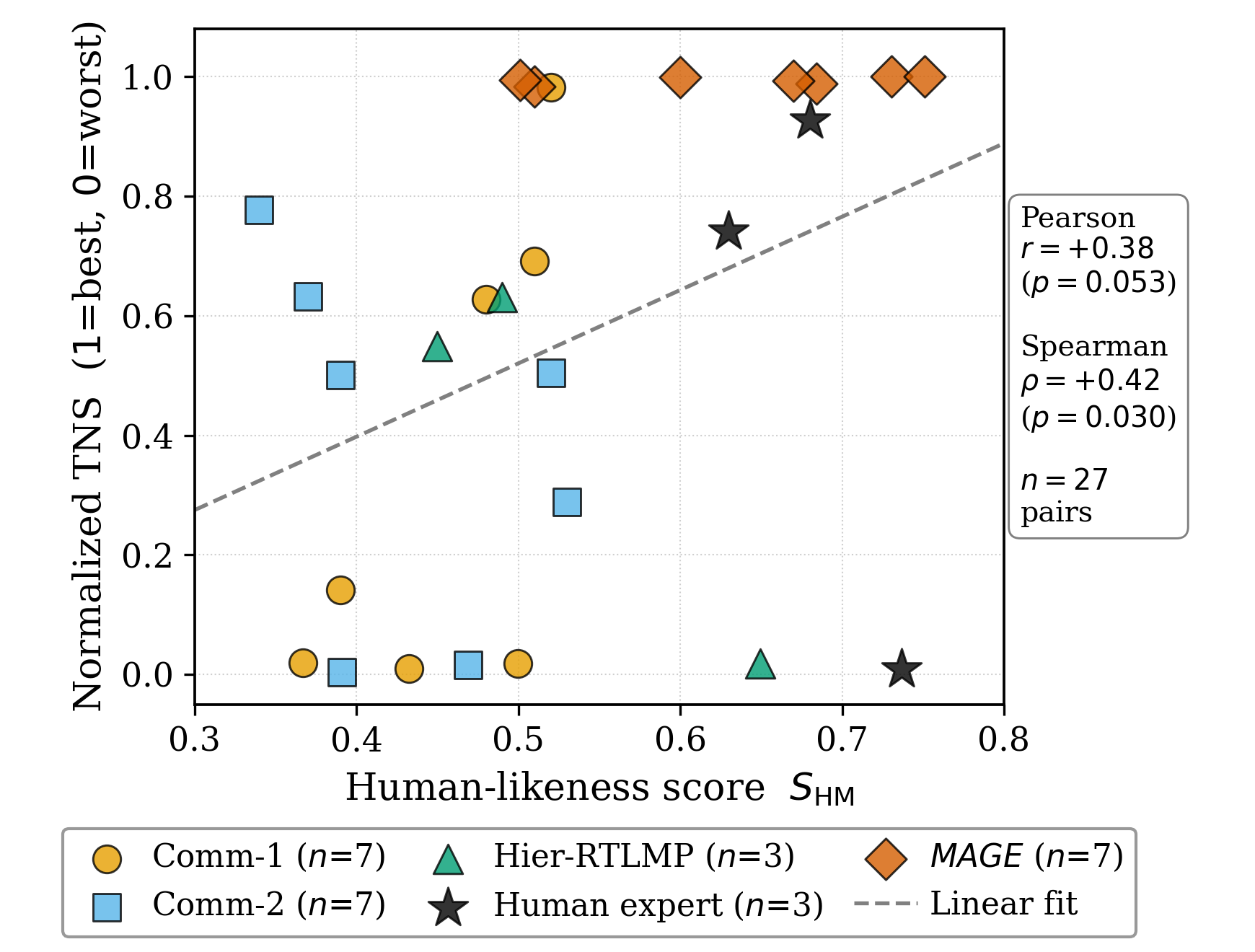}
  \caption{\textcolor{black}{Per-design min-max-normalized TNS versus human-likeness
score $S_{\mathrm{HM}}$ across all $(\text{design}, \text{method})$
pairs. For each design, TNS is min-max normalized across methods so
that $1$ denotes the best TNS and $0$ the worst. Higher human-likeness
is positively associated with better post-route timing
(Pearson $r=+0.38$, $p=0.053$; Spearman $\rho=+0.42$, $p=0.030$).}}
  \label{fig:correlation}
\end{figure}

\noindent\textbf{Timing vs.\ rWL remarks.}
We see that commercial baselines often have lower total rWL, but {\em MAGE}
still reaches better WNS and TNS. I.e., a lower total rWL does not mean
better timing.
\textcolor{black}{All methods share the same floorplan: the die and core boundary,
the IO placement, and the power-ground (PG) configuration are identical,
and only the macro placement differs. Every placement therefore has the
same total whitespace, and only locations change
(Figure~\ref{fig:placement-grid}). Comm-1 and Comm-2 pack the macros
into the interior and split the whitespace into many small pockets.
{\em MAGE} instead follows our human-like rules
(Section~\ref{sec:placement-rules}): it aligns the macros, keeps clear
macro channels, leaves an IO keep-out, and holds the whitespace in one
continuous region. We have examined the post-route paths to understand why
this helps timing. The improvement comes from three effects that act
together: clock insertion delay, clock skew, and congestion
(standard-cell density) around the critical register-to-register logic.
More concretely, for
\emph{Ariane-NG45}, the reason is the clock: {\em MAGE} lowers the worst
clock insertion delay from $0.43$ to $0.20$\,ns and the skew from $0.44$
to $0.34$\,ns, and on the same start-to-end path almost all of the slack
difference comes from clock skew, not from the datapath wiring. For
\emph{BP-NG45}, the reason is congestion: {\em MAGE} keeps the critical
logic at a lower standard-cell utilization ($45.8\%$ vs.\ $49.5\%$), so
it avoids $\sim2{,}750$ violating register-to-register paths that the
denser commercial placements add ($9{,}622$ vs.\ $6{,}868$), and this
improves TNS from $-357$ to $-139$\,ns. The trend in
Figure~\ref{fig:correlation} agrees with this. Overall, we see indications 
that the human-like placement rules are what give {\em MAGE} its better
post-route timing.}

\begin{table}[htbp]
\caption{Geometric-mean improvement of {\em MAGE} over each baseline
for human-likeness metrics. Positive values indicate improvement by
{\em MAGE}.}
\label{tab:human_geomean}
\centering
\scriptsize
\setlength{\tabcolsep}{3pt}
\renewcommand{\arraystretch}{1.1}
\begin{tabular}{@{}l rrrrr@{}}
\toprule
\textbf{{\em MAGE} vs.}
& $S_{\mathrm{notch}}$ (\%)
& $S_{\mathrm{ws}}$ (\%)
& $S_{\mathrm{pocket}}$ (\%)
& $S_{\mathrm{align}}$ (\%)
& $S_{\mathrm{HM}}$ (\%) \\
\midrule
Human      & \textbf{14.0} & \textbf{1.9}  & \textbf{4.0}  & \textbf{0.4}  & \textbf{6.0}  \\
Comm-1     & \textbf{15.2} & \textbf{7.8}  & \textbf{11.8} & \textbf{18.2} & \textbf{37.5} \\
Comm-2     & \textbf{68.1} & \textbf{12.3} & \textbf{15.6} & \textbf{15.0} & \textbf{47.8} \\
Hier-RTLMP & \textbf{32.0} & \textbf{3.6}  & \textbf{18.8} & \textbf{7.2}  & \textbf{37.4} \\
\bottomrule
\end{tabular}
\vspace{-10pt}
\end{table}

\subsection{Evaluation of Human-Likeness}
\label{sec:human_metrics}

Table~\ref{tab:human_geomean} summarizes the geometric-mean improvement of
{\em MAGE} over each baseline for the human-likeness metrics defined in
Section~\ref{sec:human_likeliness_metrics}. 
Against the human expert baseline on NG45, {\em MAGE} improves the
overall human-likeness score $S_{\mathrm{HM}}$ by 6.0\%, with the
largest gain in notch score ($+14.0\%$). \textcolor{black}{The notch
gain does not indicate that the expert placements are poorly
structured; rather, it reflects that {\em MAGE} follows 
uniform-width stacking, yielding much better aligned
macro silhouettes, whereas an expert designer tolerates minor
silhouette irregularities in exchange for other objectives such as
pin accessibility. On the remaining sub-scores
{\em MAGE} tracks the expert closely (Table~\ref{tab:human_geomean}),
showing that it can potentially match expert-level organization while
yielding better PPA.}
\textcolor{black}{Apart from the human baseline and {\em MAGE}, 
Hier-RTLMP attains the highest human-likeness, 
consistent with its design goal of producing human-quality macro 
placements~\cite{hier-rtlmp}.}
Figure~\ref{fig:correlation} shows the relationship between human-likeness and
TNS. Placements with higher $S_{\mathrm{HM}}$ tend to have better timing
outcomes.

\subsection{Ablation Studies}
\label{sec:ablation}

\textcolor{black}{We run two targeted ablations --- removing the
knowledge corpus $\mathcal{K}$ and the visual feedback pipeline
(Section~\ref{sec:vision-use}) --- on Ariane and BP (NG45). We limit
the study to two designs given the high per-run cost
(Section~\ref{sec:case_studies}).} 
For each ablation, we remove one component and
report the average change relative to the full framework.
For rWL and power, degradation is computed as
$100\cdot(M_{\mathrm{abl}}-M_{\mathrm{full}})/M_{\mathrm{full}}$, where
$M_{\mathrm{abl}}$ and $M_{\mathrm{full}}$ denote the metric value for the
ablated and full {\em MAGE} runs, respectively. 
For $S_{\mathrm{HM}}$, degradation is
computed as
$100\cdot(S_{\mathrm{full}}-S_{\mathrm{abl}})/S_{\mathrm{full}}$, where
$S_{\mathrm{abl}}$ and $S_{\mathrm{full}}$ denote the human-likeness scores of
the ablated and full runs, respectively.
\blue{Tables~\ref{tab:ablation_knowledge} and~\ref{tab:ablation_vision}
list the per-design values for the full and ablated runs; we report the
corresponding average degradation over Ariane and BP for each metric in
the discussion below.}

\noindent{\bf Effect of knowledge corpus.}
\blue{We compare the full {\em MAGE} framework against a variant
{\em MAGE}$\setminus\mathcal{K}$ in which the knowledge corpus is
removed: sub-agents~1, 3, and~5 no longer receive prior placement
knowledge, known design issues, or design collateral, and rely solely
on the information extracted from the input netlist and floorplan.}

\begin{table}[t]
\centering
\caption{\blue{Ablation: effect of knowledge corpus $\mathcal{K}$ on
Ariane and BP in NG45. Best values per design in \textbf{bold}.}}
\label{tab:ablation_knowledge}
\footnotesize
\setlength{\tabcolsep}{3pt}
\renewcommand{\arraystretch}{1.05}
\begin{tabular}{@{}ll rrrr r@{}}
\toprule
\textbf{Design} & \textbf{Variant} & \textbf{rWL} & \textbf{Pwr}
  & \textbf{WNS} & \textbf{TNS} & $S_{\mathrm{HM}}$ \\
 & & \scriptsize(mm) & \scriptsize(mW) & \scriptsize(ps)
  & \scriptsize(ns) & \\
\midrule
\multirow{2}{*}{Ariane}
 & {\em MAGE}                       & \textbf{4424}  & \textbf{832} & \textbf{$-$81}  & \textbf{$-$32.7} & \textbf{0.72} \\
 & {\em MAGE}$\setminus\mathcal{K}$ & 4536           & 836           & $-$99            & $-$60.2           & 0.73 \\
\midrule
\multirow{2}{*}{BP}
 & {\em MAGE}                       & \textbf{26987} & \textbf{4479} & \textbf{$-$88}  & \textbf{$-$139.7} & 0.68 \\
 & {\em MAGE}$\setminus\mathcal{K}$ & 28056          & 4492           & $-$132           & $-$258.0           & 0.68 \\
\bottomrule
\end{tabular}
\end{table}

\textcolor{black}{\blue{As Table~\ref{tab:ablation_knowledge} shows,
removing $\mathcal{K}$ degrades WNS by 35.2\% and TNS by 84.4\% on
average}, with little effect on rWL (3.7\%), power (0.3\%),
or $S_{\mathrm{HM}}$ ($-0.7\%$) --- showing that $\mathcal{K}$
affects timing, not structure (the principles of
Section~\ref{sec:placement-rules} are enforced via prompts regardless
of $\mathcal{K}$). \textcolor{black}{Concretely, $\mathcal{K}$ stores
prior-run artifacts (decision records, literature, accepted
macro placement patterns) indexed by design; the relevant entries are
retrieved during Phases~1, 3, and~5 and appended to the sub-agent's
prompt. For BP this supplies cues such as a preferred four-tile
quadrant symmetry, which generic principles alone cannot provide.}}



\noindent{\bf Effect of visual feedback.}
\blue{We compare the full {\em MAGE} framework against a variant
{\em MAGE}$\setminus V$ in which all visual feedback is disabled: the
mandatory visualization gates are bypassed, the auxiliary vision
pipeline (Section~\ref{sec:aux-vision}) is removed, and the three
visual quality checks are skipped, so all spatial reasoning must be
performed from numerical data alone.}

\begin{table}[t]
\centering
\caption{\blue{Ablation: effect of visual feedback on Ariane and BP in
NG45. Best values per design in \textbf{bold}.}}
\label{tab:ablation_vision}
\footnotesize
\setlength{\tabcolsep}{3pt}
\renewcommand{\arraystretch}{1.05}
\begin{tabular}{@{}ll rrrr r@{}}
\toprule
\textbf{Design} & \textbf{Variant} & \textbf{rWL} & \textbf{Pwr}
  & \textbf{WNS} & \textbf{TNS} & $S_{\mathrm{HM}}$ \\
 & & \scriptsize(mm) & \scriptsize(mW) & \scriptsize(ps)
  & \scriptsize(ns) & \\
\midrule
\multirow{2}{*}{Ariane}
 & {\em MAGE}              & \textbf{4424}  & \textbf{832}  & \textbf{$-$81}  & \textbf{$-$32.7}  & \textbf{0.72} \\
 & {\em MAGE}$\setminus V$ & 4628           & 839            & $-$119           & $-$109.0           & 0.70 \\
\midrule
\multirow{2}{*}{BP}
 & {\em MAGE}              & \textbf{26987} & 4479           & \textbf{$-$88}  & \textbf{$-$139.7} & \textbf{0.68} \\
 & {\em MAGE}$\setminus V$ & 27145          & \textbf{4475}  & $-$122           & $-$221.9           & 0.64 \\
\bottomrule
\end{tabular}
\vspace{-5pt}
\end{table}

\begin{figure}[htpb]
\centering
\includegraphics[width=0.48\columnwidth]{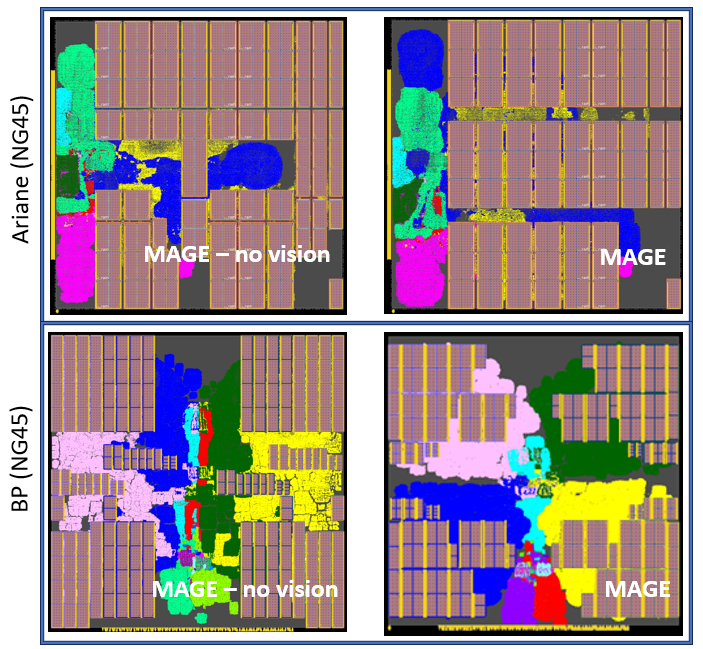}
\caption{Effect of visual feedback on Ariane and BP (NG45). Without visual
feedback, macro groups become fragmented; full {\em MAGE} produces more
human-like macro placements.}
\label{fig:vision_ablation}
\end{figure}

\textcolor{black}{\blue{As Table~\ref{tab:ablation_vision} shows,
removing visual feedback degrades WNS by 40.5\% and TNS by 133.0\%},
with $S_{\mathrm{HM}}$ dropping 4.2\%, while rWL (2.3\%)
and power (0.4\%) are largely unaffected. Figure~\ref{fig:vision_ablation}
shows the reason: without vision, macros scatter with irregular gaps
and poor boundary use, whereas the full framework snaps them to the
boundary with structured stacks. These are issues visual inspection catches
but numerical checks miss.}


\subsection{Case studies}
\label{sec:case_studies}

We present four case studies that evaluate {\em MAGE} beyond the standard
benchmark setting: anonymized netlists, unseen designs, rectilinear floorplans,
and high-utilization layouts. These studies test whether {\em MAGE} depends on
semantic design information, generalizes zero-shot, and remains effective on
non-rectangular or area-constrained floorplans.

\begin{table}[htbp]
\centering
\caption{\textcolor{black}{Effect of netlist anonymization on {\em MAGE}. ``Anon.''
denotes the anonymized variant. Best results in \textbf{bold}.}}
\label{tab:anonymized}
\scriptsize
\setlength{\tabcolsep}{3pt}
\renewcommand{\arraystretch}{1.05}
\begin{tabular}{@{}ll rrrr@{}}
\toprule
\textbf{Design (NG45)} & \textbf{Variant} & \textbf{rWL (mm)} & \textbf{Pwr (mW)}
  & \textbf{WNS (ps)} & \textbf{TNS (ns)} \\
\midrule
\multirow{2}{*}{Ariane}
 & {\em MAGE} (Anon.) & 4687           & 841           & -86           & -69.9 \\
 & {\em MAGE}         & \textbf{4424}  & \textbf{832}  & \textbf{-81}  & \textbf{-32.7} \\
\midrule
\multirow{2}{*}{BP}
 & {\em MAGE} (Anon.) & \textbf{25766} & \textbf{4466} & -126          & -310.8 \\
 & {\em MAGE}         & 26987          & 4479          & \textbf{-88}  & \textbf{-139.7} \\
\bottomrule
\end{tabular}
\end{table}

\noindent\textbf{1. Anonymized designs.}
To evaluate the role of semantic design information, we anonymize Ariane
and BP by replacing module, instance, net, and port names with generic
identifiers while preserving hierarchy, connectivity, macro dimensions,
and floorplan geometry. This also disables the knowledge corpus
$\mathcal{K}$, since design-specific entries can no longer be retrieved.
Table~\ref{tab:anonymized} shows that anonymization degrades timing: 
TNS worsens from
$-32.7$\,ns to $-69.9$\,ns on Ariane and from $-139.7$\,ns to
$-310.8$\,ns on BP. \textcolor{black}{Because
connectivity and geometry are preserved, this case study
isolates what names contribute \emph{beyond} structure: descriptive
identifiers (e.g., \texttt{tile\_y0x0}, \texttt{pu\_core[2]}) let
the LLM recognize regular arrays, functional groupings, and symmetry
that are not recoverable from the raw connectivity matrix.
When names are stripped, the framework must infer these relationships
from connectivity alone, which is harder and yields less
symmetrical group placement.}

\begin{table}[htpb]
\centering
\caption{Results on unseen designs (zero-shot generalization). Below each
design name, the triplet (Inst., Macros, Types) reports the total instance
count, the number of hard macros, and the number of unique macro types,
respectively. All designs are GF12, so values are normalized as in
Table~\ref{tab:results_ppa}: rWL and Pwr to Comm-1, WNS and TNS to the
target clock period. The best value per design is shown in
\textbf{bold}.}
\label{tab:unseen}
\footnotesize
\setlength{\tabcolsep}{3pt}
\renewcommand{\arraystretch}{1.05}
\begin{tabular}{@{}ll rrrr r@{}}
\toprule
\textbf{Design} & \textbf{Method} & \textbf{rWL} & \textbf{Pwr}
  & \textbf{WNS} & \textbf{TNS} & $S_{\mathrm{HM}}$ \\
\midrule
\multirow{2}{*}{\shortstack[l]{VTA\\\scriptsize(168K, 83, 8)}}
 & Comm-1      & 1.000 & 1.000 & -0.22 & -170.4 & 0.5 \\
 & {\em MAGE}  & 1.142 & 1.022 & \textbf{-0.16} & \textbf{-101.8} & \textbf{0.7} \\
\midrule
\multirow{2}{*}{\shortstack[l]{GeneSys01\\\scriptsize(695K, 368, 6)}}
 & Comm-1      & 1.000 & 1.000 & -0.17 & -82.7 & 0.5 \\
 & {\em MAGE}  & 1.157 & 1.009 & \textbf{-0.16} & \textbf{-60.7} & \textbf{0.6} \\
\bottomrule
\end{tabular}
\end{table}

\begin{figure}[htpb]
\centering
\includegraphics[width=0.48\columnwidth]{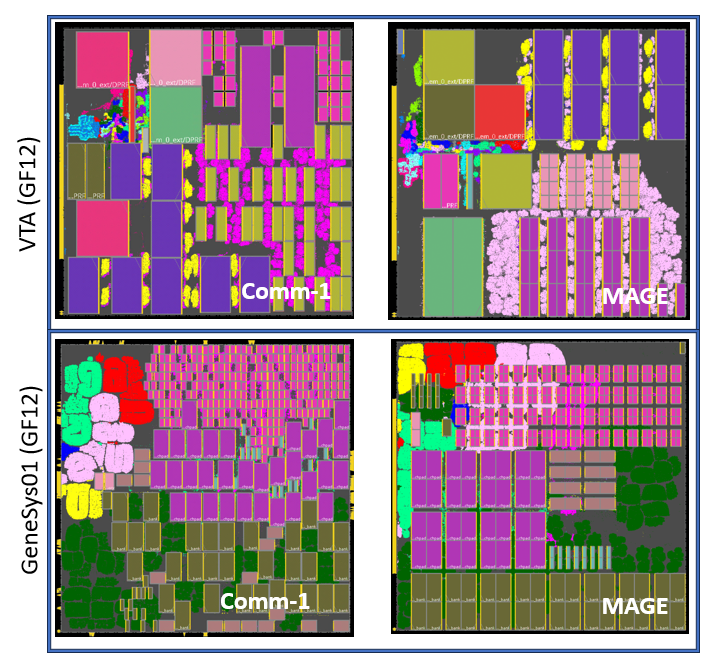}
\caption{Visual comparison of Comm-1 and {\em MAGE} on two unseen designs. 
\textbf{Top:} \textsc{VTA} (GF12). \textbf{Bottom:}
\textsc{GeneSys01} (GF12). Left: Comm-1.
Right: {\em MAGE}.}
\label{fig:vision_unseen}
\end{figure}

\noindent \textbf{2. Unseen designs.}
To test zero-shot generalization, we apply {\em MAGE} to two GF12
designs, VTA and GeneSys01, that were not used during prompt
development or knowledge-corpus construction. As shown in
Table~\ref{tab:unseen}, {\em MAGE} improves timing over Comm-1 on both
designs: WNS/TNS improve by 27\%/40\% on VTA and by 6\%/27\% on
GeneSys01. {\em MAGE} also achieves higher human-likeness scores
($0.70$ vs.\ $0.50$ on VTA and $0.63$ vs.\ $0.50$ on GeneSys01), at the
cost of 14\%--16\% higher rWL. Figure~\ref{fig:vision_unseen}
illustrates the same pattern observed on the main benchmarks: {\em MAGE}
forms cleaner boundary-aligned stacks and more contiguous whitespace,
while Comm-1 produces \textcolor{black}{less human-like macro placements}.

\begin{figure}[htbp]
\centering
\begin{subfigure}[t]{0.45\columnwidth}
    \centering
    \includegraphics[width=\linewidth]{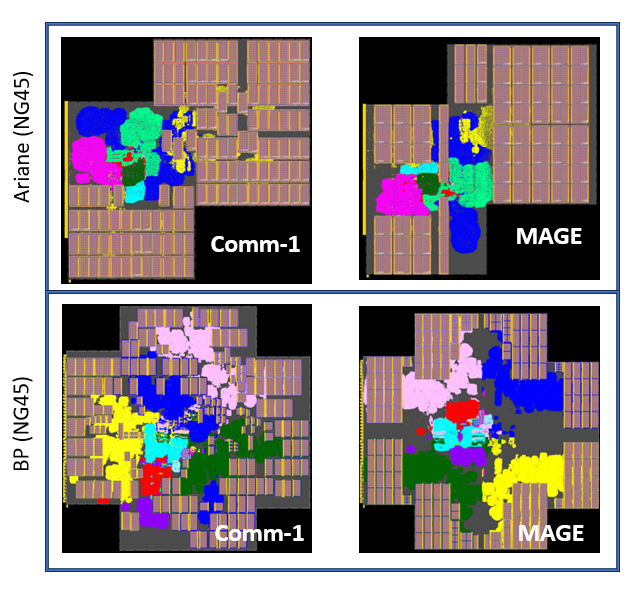}
    \caption{Hor. and ver. convexity.}
    \label{fig:cutout_hv_convex}
\end{subfigure}
\hfill
\begin{subfigure}[t]{0.45\columnwidth}
    \centering
    \includegraphics[width=\linewidth]{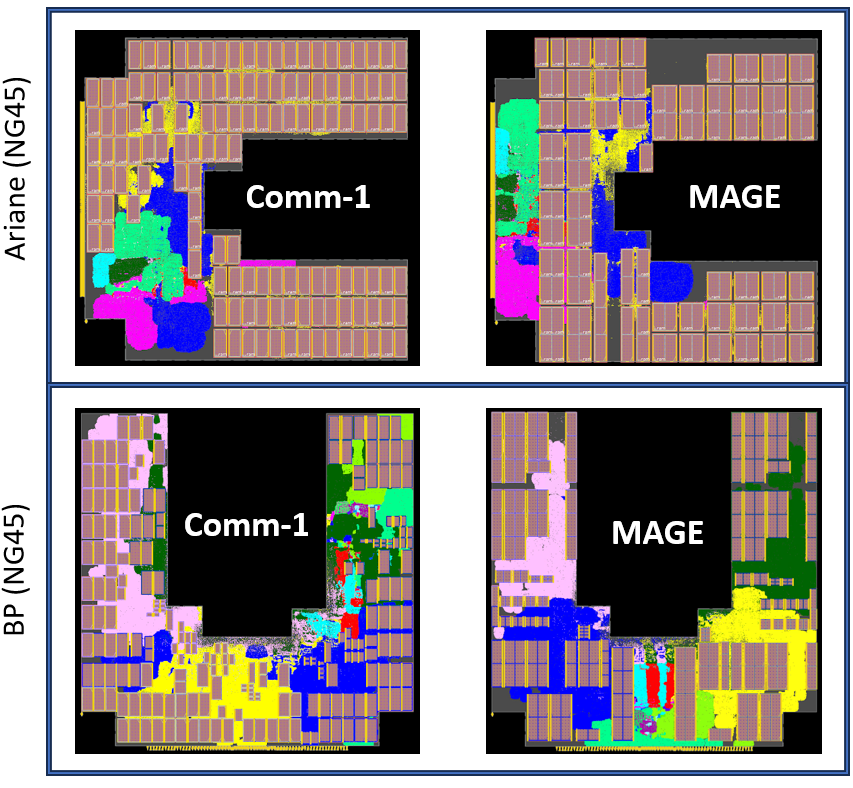}
    \caption{Mixed convexity.}
    \label{fig:cutout_mixed_convex}
\end{subfigure}

\caption{Placement comparison on dense rectilinear floorplans
($>$70\% utilization). In both subfigures, \textbf{top:} \textsc{Ariane}
(NG45), \textbf{bottom:} \textsc{BP} (NG45), \textbf{left:} Comm-1, and
\textbf{right:} {\em MAGE}.}
\label{fig:cutout_floorplans_combined}
\end{figure}



\begin{table}[htpb]
\centering
\caption{Results on dense rectilinear floorplan designs ($>$70\%
utilization). HV: horizontal and vertical convexity
(Figure~\ref{fig:cutout_hv_convex}). Mixed: horizontal top,
vertical bottom (Figure~\ref{fig:cutout_mixed_convex}). Best
values per design-variant in \textbf{bold}.\protect\footnotemark}
\label{tab:rectilinear_results}
\scriptsize
\setlength{\tabcolsep}{3pt}
\renewcommand{\arraystretch}{1.05}
\begin{tabular}{@{}lll rrrr@{}}
\toprule
\textbf{Design (NG45)} & \textbf{Type} & \textbf{Method}
  & \textbf{rWL (mm)} & \textbf{Pwr (mW)} & \textbf{WNS (ps)} & \textbf{TNS (ns)} \\
\midrule
\multirow{4}{*}{Ariane}
 & \multirow{2}{*}{HV}
   & Comm-1     & \textbf{3981}  & \textbf{829}  & -117          & -76.6  \\
 & & {\em MAGE} & 4859           & 829           & \textbf{-105} & \textbf{-70.1} \\
\cmidrule(l){2-7}
 & \multirow{2}{*}{Mixed}
   & Comm-1     & \textbf{4273}  & \textbf{835}  & -118          & \textbf{-92.4} \\
 & & {\em MAGE} & 4934           & 845           & \textbf{-112} & -94.0 \\
\midrule
\multirow{4}{*}{BP}
 & \multirow{2}{*}{HV}
   & Comm-1     & \textbf{24072} & \textbf{4450} & -205          & -1250.1 \\
 & & {\em MAGE} & 32757          & 4597          & \textbf{-123} & \textbf{-326.9} \\
\cmidrule(l){2-7}
 & \multirow{2}{*}{Mixed}
   & Comm-1     & 32566          & 4652          & -415          & \textbf{-11667.1} \\
 & & {\em MAGE} & \textbf{32252} & \textbf{4633} & \textbf{-354} & -11959.2 \\
\bottomrule
\end{tabular}
\vspace{-10pt}
\end{table}
\footnotetext{\textcolor{black}{Convexity refers to the shape of the
rectilinear core cutout. In an \emph{HV} (horizontal-and-vertical
convex) variant, the core is clipped along both a horizontal and a
vertical edge, producing an L-shaped boundary that is convex in both
directions. In a \emph{Mixed} variant, the cutout is horizontal along
the top and vertical along the bottom, yielding a boundary whose
convex direction differs between its upper and lower halves.}}

\noindent \textbf{3. Rectilinear floorplan designs.}
We next evaluate dense rectilinear floorplans for Ariane and BP in NG45,
with utilization above 70\%. Table~\ref{tab:rectilinear_results} shows
that {\em MAGE} improves WNS on all four rectilinear variants. The
largest gain occurs on BP-HV, where WNS improves from $-205$\,ps to
$-123$\,ps and TNS improves from $-1250.1$\,ns to $-326.9$\,ns. The
mixed-convexity cases show some tradeoff: {\em MAGE} improves
WNS but does not always improve TNS. Overall, these results show
that {\em MAGE} generalizes to rectilinear floorplan designs.

\begin{figure}[t]
\centering
\includegraphics[width=\columnwidth]{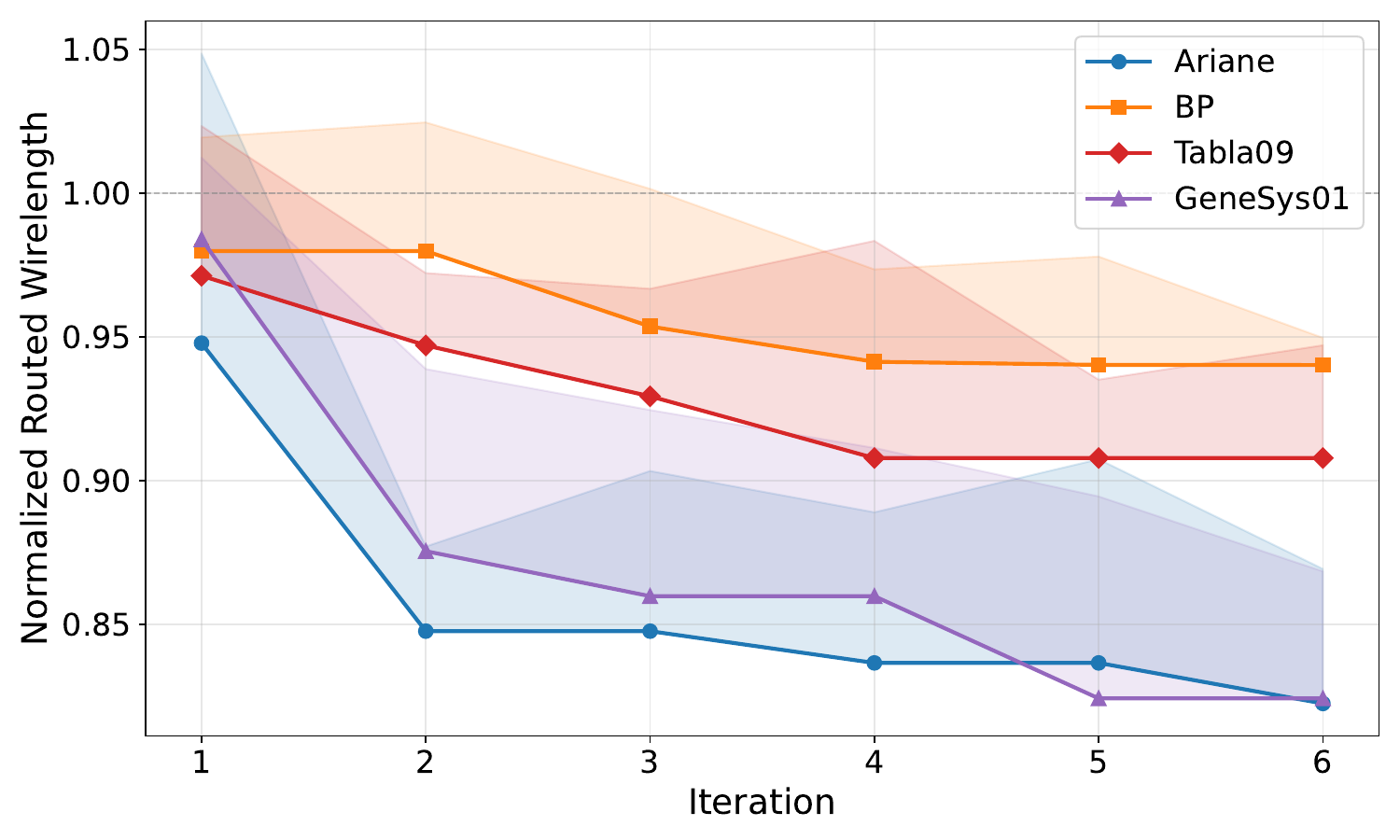}
\caption{Normalized routed wirelength across tournament iterations for
four designs (normalized to iteration~1 mean). Solid lines: best
variant per iteration. Shaded regions: spread between worst and best
variants. The monotonically decreasing trajectories demonstrate the
effectiveness of the analysis and reflection agents in guiding
iterative improvement.}
\label{fig:evolution}
\end{figure}

\noindent
\textbf{Evolution trajectory.}
Figure~\ref{fig:evolution} plots the normalized routed wirelength of
{\em MAGE}'s variants across tournament iterations for four designs:
Ariane, BP, Tabla09, and Genesys01. Values are normalized to the
iteration~1 mean, and the shaded region shows the spread between worst
and best variants in each round.
All four designs exhibit a downward trajectory, confirming that the
tournament's analysis and reflection agents
(Section~\ref{sec:tournament}) successfully guide iterative improvement.
Ariane shows the strongest convergence, with the best variant dropping
from 0.95$\times$ to 0.83$\times$ of the initial mean (a $\sim$13\%
improvement) by iteration~6, accompanied by a narrowing spread that
indicates the variant pool concentrating around stronger solutions. BP
converges more gradually, reaching $\sim$0.91$\times$ by iteration~6
($\sim$7\% improvement), with a wider shaded region reflecting the
greater difficulty of placing 220 macros across 6 types. Tabla09 shows
steady improvement from 0.98$\times$ to $\sim$0.90$\times$ ($\sim$8\%
improvement), with the tightest spread among all designs, suggesting
that the smaller design (368 macros) leaves less room for variant
diversity. Genesys01 (an unseen design) shows a trajectory comparable
to the main benchmarks, reaching $\sim$0.91$\times$ by iteration~6,
confirming that the tournament mechanism generalizes to designs not used
during prompt development.
Across all four designs, the tournament yields an average 9\% rWL
improvement over the iteration~1 mean. Most of the gain occurs in the
first 2--3 iterations, after which the trajectory plateaus---suggesting
that $R = 3$ could suffice in practice.

\begin{figure}[htpb]
\centering
\scriptsize
\setlength{\tabcolsep}{3pt}
\renewcommand{\arraystretch}{1.0}
\begin{tabular}{@{}c c @{\hspace{6pt}} c @{\hspace{6pt}} c @{\hspace{6pt}} c@{}}
 & & 68\% & 75\% & 80\% \\[2pt]
\multirow{2}{*}{\rotatebox{90}{Ariane}}
 & \rotatebox{90}{Comm-1} &
\includegraphics[width=0.078\textwidth]{Figures/Ariane-GF12-CMP.png} &
\includegraphics[width=0.078\textwidth]{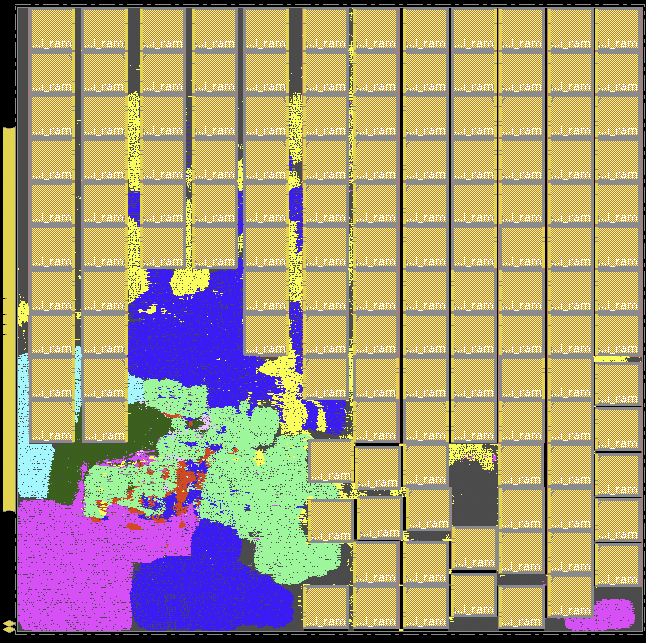} &
\includegraphics[width=0.078\textwidth]{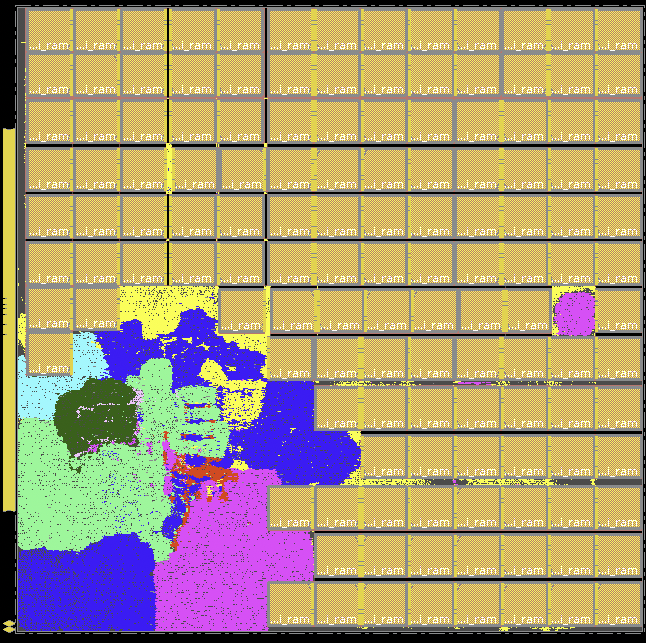} \\[2pt]
 & \rotatebox{90}{\em MAGE} &
\includegraphics[width=0.078\textwidth]{Figures/Ariane-GF12-MAGE.png} &
\includegraphics[width=0.078\textwidth]{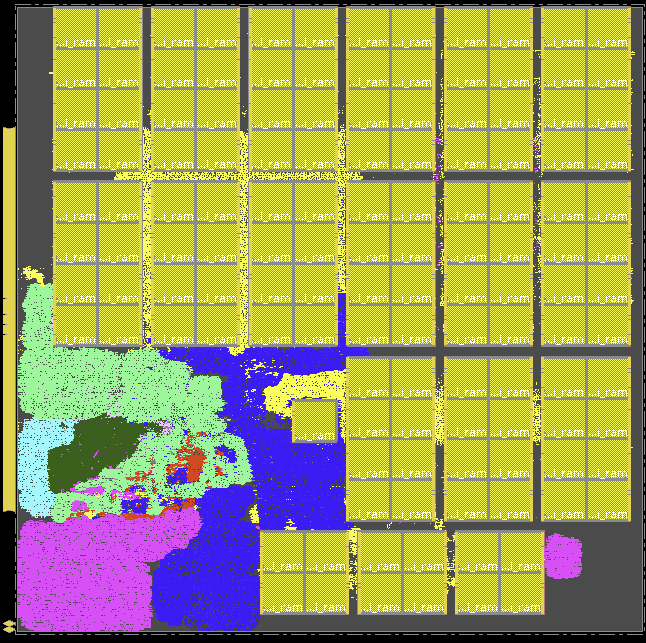} &
\includegraphics[width=0.078\textwidth]{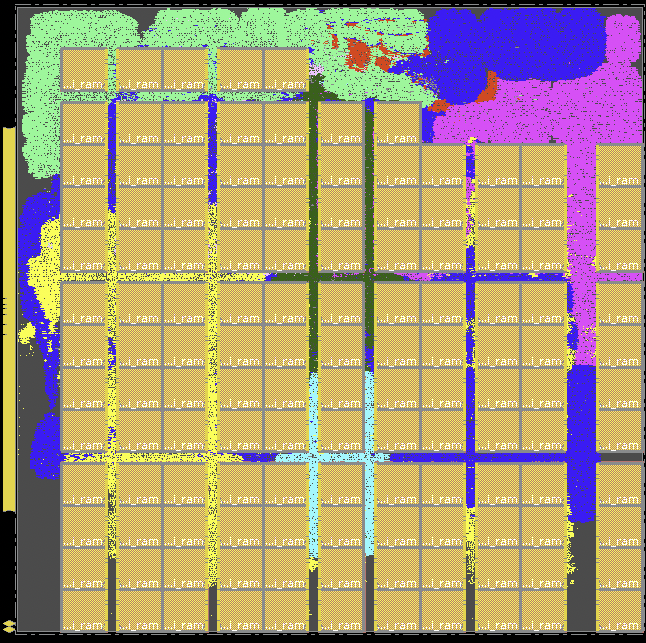} \\[4pt]
\multirow{2}{*}{\rotatebox{90}{BP}}
 & \rotatebox{90}{Comm-1} &
\includegraphics[width=0.078\textwidth]{Figures/BP-GF12-CMP.png} &
\includegraphics[width=0.078\textwidth]{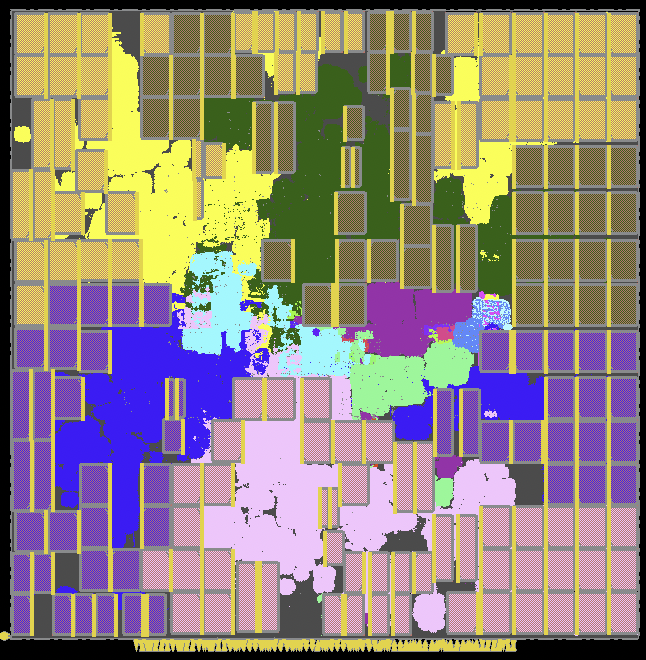} &
\includegraphics[width=0.078\textwidth]{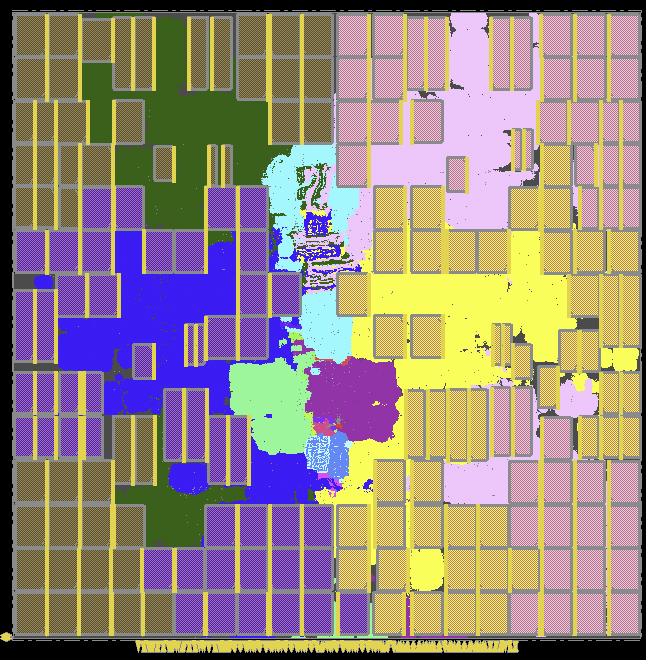} \\[2pt]
 & \rotatebox{90}{\em MAGE} &
\includegraphics[width=0.078\textwidth]{Figures/BP-GF12-MAGE.png} &
\includegraphics[width=0.078\textwidth]{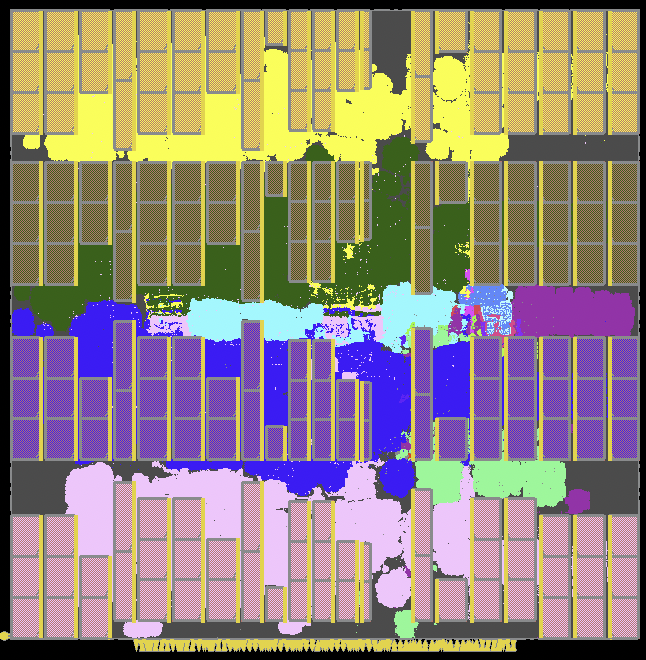} &
\includegraphics[width=0.078\textwidth]{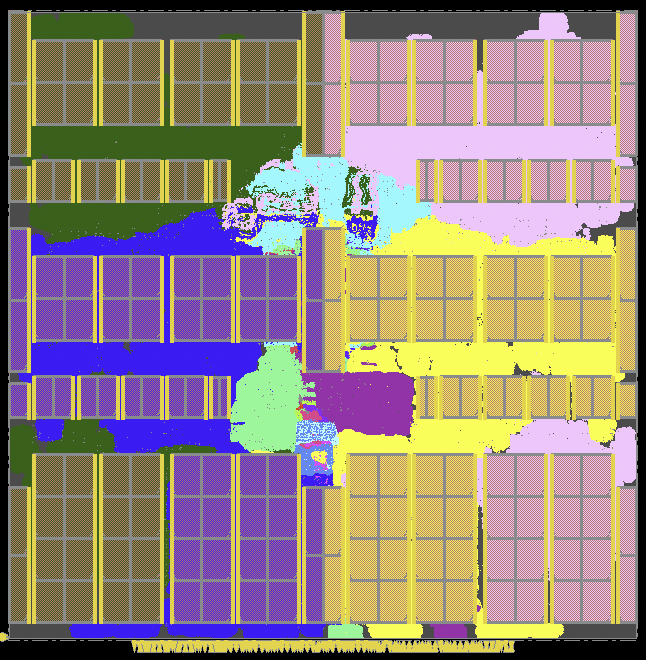} \\
\end{tabular}
\caption{Placement comparison across utilization levels. {\em MAGE}
preserves regular stacking as utilization increases, while Comm-1
placements become more fragmented.}
\label{fig:util_sweep}
\end{figure}

\begin{table}[htpb]
\centering
\caption{Results with utilization sweep. Enablement: GF12. Values are
normalized: rWL and Pwr to the Comm-1 result at each utilization, and WNS
and TNS to the target clock period (fixed across utilizations). Best
values per design--utilization pair in \textbf{bold}.}
\label{tab:util_results}
\scriptsize
\setlength{\tabcolsep}{3pt}
\renewcommand{\arraystretch}{1.05}
\begin{tabular}{@{}lll rrrr@{}}
\toprule
\textbf{Design} & \textbf{Util} & \textbf{Method}
  & \textbf{rWL} & \textbf{Pwr} & \textbf{WNS} & \textbf{TNS} \\
\midrule
\multirow{6}{*}{Ariane}
 & \multirow{2}{*}{68}
   & Comm-1       & \textbf{1.000} & \textbf{1.000} & -0.16 & -142.3 \\
 & & {\em MAGE}   & 1.223 & 1.021 & \textbf{-0.10}  & \textbf{-68.0}  \\
\cmidrule(l){2-7}
 & \multirow{2}{*}{75}
   & Comm-1       & 1.000 & 1.000 & -0.16 & -130.9 \\
 & & {\em MAGE}   & \textbf{0.964} & \textbf{0.988} & \textbf{-0.10}  & \textbf{-69.9}  \\
\cmidrule(l){2-7}
 & \multirow{2}{*}{80}
   & Comm-1       & \textbf{1.000} & \textbf{1.000} & -0.13 & -137.4 \\
 & & {\em MAGE}   & 1.259 & 1.037 & \textbf{-0.11}  & \textbf{-107.8}  \\
\midrule
\multirow{6}{*}{BP}
 & \multirow{2}{*}{68}
   & Comm-1       & \textbf{1.000} & \textbf{1.000} & -0.09 & -138.9 \\
 & & {\em MAGE}   & 1.099 & 1.010 & \textbf{-0.08} & \textbf{-33.4}  \\
\cmidrule(l){2-7}
 & \multirow{2}{*}{75}
   & Comm-1       & \textbf{1.000} & \textbf{1.000} & -0.11 & -347.3 \\
 & & {\em MAGE}   & 1.081 & 1.004 & \textbf{-0.05} & \textbf{-29.6} \\
\cmidrule(l){2-7}
 & \multirow{2}{*}{80}
   & Comm-1       & \textbf{1.000} & 1.000 & -0.33 & -2850.0 \\
 & & {\em MAGE}   & 1.105 & \textbf{0.987} & \textbf{-0.07}  & \textbf{-42.9}  \\
\bottomrule
\end{tabular}
\vspace{-10pt}
\end{table}

\noindent
\textbf{4. Utilization sweep.}
\textcolor{black}{Last, to evaluate {\em MAGE}} under tighter area constraints, we 
increase utilization on Ariane and BP in GF12 while keeping the
target clock period fixed. In this sweep, rWL and Pwr are normalized to
the Comm-1 result at each utilization, while WNS and TNS are normalized to
the target clock period. Table~\ref{tab:util_results} shows that
{\em MAGE} achieves the best WNS and TNS across all six design--utilization
pairs. The advantage is largest at high utilization. On BP at 80\%
utilization, {\em MAGE} improves normalized WNS from $-0.33$ to $-0.07$ and
normalized TNS from $-2850.0$ to $-42.9$. Figure~\ref{fig:util_sweep} shows \textcolor{black}{that}
as utilization increases, Comm-1 placements become more fragmented, while
{\em MAGE} preserves boundary-snapped stacks and organized routing channels.
These results \textcolor{black}{show} that the placement principles used by {\em MAGE}
remain effective under tighter area constraints.

\noindent\textbf{LLM resource usage.}
Each {\em MAGE} tournament run uses 6 variants across 6 rounds. For
a single-variant run, input tokens range
from 78M to 189M and output tokens from 1.8M to 4.3M, at a \textcolor{black}{current} 
cost of \$6--\$14 per variant. A full tournament run costs \$220--\$509, averaging
\$310 per design. Wall-clock tournament runtime ranges from 4.9 hours to
18.8 hours, averaging 10.0 hours; variants within each round execute
in parallel, so runtime tracks the slowest variant per round. \textcolor{black}{Unsurprisingly, the} 
runtime is
dominated not by LLM generation but by downstream P\&R evaluation\textcolor{black}{, e.g.,} MemPool's 2.6M instances \textcolor{black}{induce long} P\&R runs while token usage \textcolor{black}{is} comparable to the other designs. \textcolor{black}{We emphasize that while}
{\em MAGE} is slower than conventional macro placers \textcolor{black}{which might run in seconds to minutes},
it targets a different operating point: producing human-like macro
placements to reduce or eliminate manual \textcolor{black}{refinement steps that today require days to weeks} of engineer time.


\section{Conclusion}
\label{sec:conclusion}

We have presented {\em MAGE}, a multimodal multi-agent framework for
human-like macro placement. Rather than treating macro placement as a
single monolithic optimization problem, {\em MAGE} decomposes the task into
a six-phase workflow with explicit visual feedback and validation. Expert
floorplanning knowledge is encoded through structured natural-language
directives, visual checks, and refinement criteria, allowing the framework to
reason about hierarchy, grouping, boundary \textcolor{black}{placement}, stacking regularity,
whitespace quality, and pin accessibility without per-design training.
Across NG45 and GF12 designs, {\em MAGE} improves post-route timing over
academic and commercial baselines, with the largest gains \textcolor{black}{seen} 
in TNS. It also
matches or exceeds the human expert baseline in structural organization while
maintaining comparable wirelength and power. \textcolor{black}{Human-likeness} metrics,
ablation studies, and case studies on anonymized, unseen, rectilinear, and
high-utilization designs \textcolor{black}{all} support a central premise \textcolor{black}{that}: 
human-like
macro organization provides a useful structural prior for downstream routing
and timing closure.
More broadly, {\em MAGE} shows that multimodal reasoning can encode expert
physical-design judgment in a transparent and editable form. Future work
includes reducing wirelength and power overhead through stronger PPA-aware
feedback, extending the framework to soft macros and standard-cell-aware
floorplanning, and applying the same methodology to adjacent floorplanning
tasks such as pin access planning, power-grid planning, and \textcolor{black}{stacked
multi-die (3DIC)} floorplanning.

\noindent
\textbf{Acknowledgments. } We acknowledge the use of Claude Opus 4.6 to assist in implementing
{\em MAGE}. The final code was thoroughly reviewed, verified, and edited by us. We maintain full accountability for the content and accuracy of the codebase. No AI was used for generating any text or images in the manuscript.

\balance

\begin{IEEEbiography}
[{\raisebox{0.35in}{\includegraphics[height=1.0in, clip, keepaspectratio]
{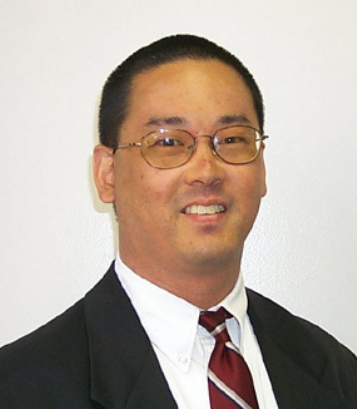}}}]
{Andrew B. Kahng} is Distinguished Professor of CSE and ECE at the 
University of California, San Diego. His interests include IC physical design, 
the design-manufacturing interface, combinatorial optimization, 
and AI/ML for EDA and IC design. He received the Ph.D. degree in Computer Science from the University of California, San Diego.
\end{IEEEbiography}
\vspace{-0.4in}

\begin{IEEEbiography}
[{\raisebox{0.6in}{\includegraphics[height=1.0in, clip, keepaspectratio]{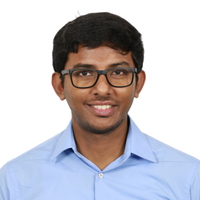}}}]
{Sayak Kundu} is a Ph.D. student in the ECE
department at the University of California,
San Diego. He received his bachelor’s degree in Electronics and Telecommunication Engineering from Jadavpur University, Kolkata, in 2017.
His research interests include optimization and machine learning applications in IC physical design flow and methodology.
\end{IEEEbiography}
\vspace{-0.4in}

\begin{IEEEbiography}
[{\raisebox{0.6in}{\includegraphics[width=0.9in, height=1in, clip,keepaspectratio]
{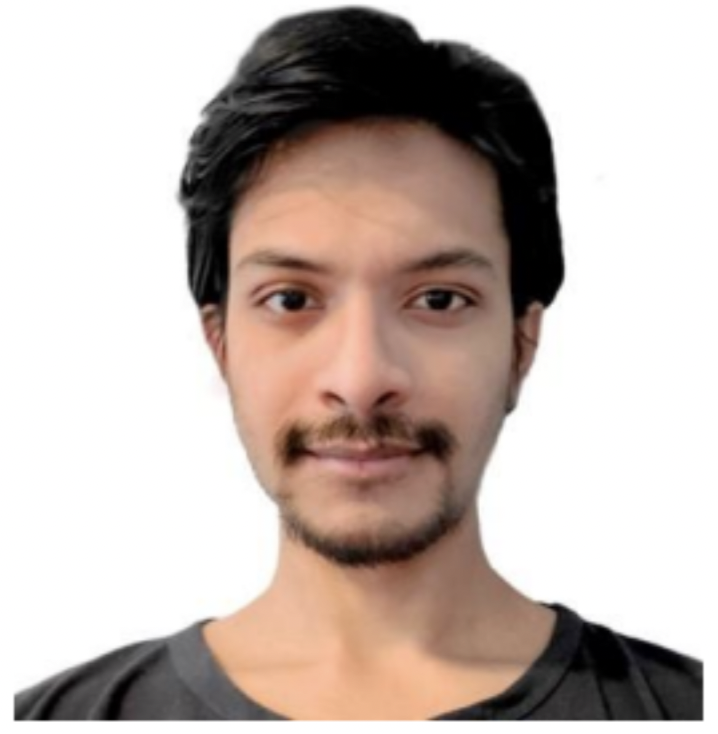}}}]
{Bodhisatta Pramanik}
received the M.S. degree in computer
engineering from the Iowa State University, Iowa, 
in 2022. He is currently pursuing the Ph.D. degree at the University of California 
at San Diego, La Jolla. His research interests include 
partitioning, placement, large-scale combinatorial optimization, and agentic EDA.
\end{IEEEbiography}


\begin{thebibliography}{99}

\bibitem{AdyaMarkov2003}
S.~N. Adya and I.~L. Markov,
``Fixed-outline floorplanning: Enabling hierarchical
design'',
{\em IEEE TVLSI}, 11(6) (2003), pp. 1120--1135.

\bibitem{AldousV94}
D. Aldous and U. Vazirani,
```Go with the winners' algorithms'',
{\em Proc. IEEE FOCS}, 1994,
pp. 492--501.


\bibitem{AutoDMP}
A. Agnesina, P. Rajvanshi, T. Yang, G. Pradipta, A. Jiao,
B. Keller et al.,
``AutoDMP: Automated DREAMPlace-based macro placement'',
{\em Proc. ISPD}, 2023, pp. 149--157.

\bibitem{BanerjeeBCDJK21}
S. Banerjee, S. Burns, P. Cocchini, A. Davare, S. Jain,
D. Kirkpatrick et al.,
``A highly configurable hardware/software stack for DNN inference
acceleration'',
{\em arXiv:2111.15024}, 2021.

\bibitem{Chang2000}
Y.-C. Chang, Y.-W. Chang, G.-M. Wu,
and S.-W. Wu,
``B*-trees: A new representation for non-slicing
floorplans'',
{\em Proc. DAC}, 2000, pp. 458--463.

\bibitem{Chen2008}
T.~Chen, Y.~Chang, and S.~Lin,
``A new multilevel framework for large-scale
interconnect-driven floorplanning'',
{\em IEEE TCAD}, 27(2) (2008), pp. 286--294.

\bibitem{Chen2014}
Y.-F. Chen, C.-C. Huang, C.-H. Chiou, Y.-W. Chang
and C.-J. Wang,
``Routability-driven blockage-aware macro placement'',
{\em Proc. DAC}, 2014, pp. 1--6.

\bibitem{ChengKKW18}
C.-K. Cheng, A.~B. Kahng, I. Kang and L. Wang,
``RePlAce: Advancing solution quality and routability validation
in global placement'',
{\em IEEE TCAD}, 38(9) (2018), pp. 1717--1730.

\bibitem{ChengKKWW23}
C.-K. Cheng, A.~B. Kahng, S. Kundu, Y. Wang and Z. Wang,
``Assessment of reinforcement learning for macro placement'',
{\em Proc. ISPD}, 2023, pp. 158--166.

\bibitem{ChengKKWW25}
C.-K. Cheng, A.~B. Kahng, S. Kundu, Y. Wang and Z. Wang,
``An updated assessment of reinforcement learning for macro
placement'',
{\em IEEE TCAD}, (2025) (DOI 10.1109/TCAD.2025.3644293).

\bibitem{Chiou2016}
C.-H. Chiou, C.-H. Chang, S.-T. Chen and Y.-W. Chang,
``Circular-contour-based obstacle-aware macro placement'',
{\em Proc. ASP-DAC}, 2016, pp. 172--177.

\bibitem{ChoiBazargan2003}
W.~Choi and K.~Bazargan,
``Hierarchical global floorplacement using simulated
annealing and network flow area migration'',
{\em Proc. DATE}, 2003, pp. 1104--1105.

\bibitem{Chuang2010}
Y.-L. Chuang, G.-J. Nam, C.~J. Alpert, Y.-W. Chang, J. Roy
and N. Viswanathan,
``Design-hierarchy aware mixed-size placement for routability
optimization'',
{\em Proc. ICCAD}, 2010, pp. 663--668.

\bibitem{Cong2006}
J.~Cong, M.~Romesis, and J.~R. Shinnerl,
``Fast floorplanning by lookahead enabled recursive
bipartitioning'',
{\em IEEE TCAD}, 25(9) (2006), pp. 1719--1732.

\bibitem{Ekpanyapong2006}
M.~Ekpanyapong, J. Minz, T. Watewai, H.-H.~S. Lee
and S.~K. Lim,
``Profile-guided microarchitectural floorplanning for deep
submicron processor design'',
{\em IEEE TCAD}, 25(7) (2006), pp. 1289--1300.

\bibitem{EsmaeilzadehGGGKK21}
H. Esmaeilzadeh, S. Ghodrati, J. Gu, S. Guo, A.~B. Kahng,
J.~K. Kim et al.,
``VeriGOOD-ML: An open-source flow for automated ML hardware
synthesis'',
{\em Proc. ICCAD}, 2021, pp. 1--7.

\bibitem{GhoseJKL26}
A. Ghose, J. Jang, A.~B. Kahng and J. Lee,
``Automated QoR improvement in OpenROAD with coding agents'',
{\em arXiv:2601.06268}, 2026.

\bibitem{Gwee1999}
B.~H. Gwee and M.~H. Lim,
``A GA with heuristic-based decoder for IC
floorplanning'',
{\em Integration}, 28(2) (1999), pp. 157--172.

\bibitem{He2020}
Z.~He, Y.~Ma, L.~Zhang, and X.~Zhou,
``Learn to floorplan through acquisition of
effective local search heuristics'',
{\em Proc. ICCD}, 2020, pp. 324--331.

\bibitem{Hsu2014}
M.-K. Hsu, Y.-F. Chen, C.-C. Huang, S. Chou, T.-H. Lin,
T.-C. Chen et al.,
``NTUplace4h: A novel routability-driven placement algorithm for
hierarchical mixed-size circuit designs'',
{\em IEEE TCAD}, 33(12) (2014), pp. 1914--1927.

\bibitem{Hu2004}
C.-C. Hu, D.-S. Chen, and Y.-W. Wang,
``Fast multilevel floorplanning for large scale
modules'',
{\em Proc. ISCAS}, 2004, pp. 205--208.

\bibitem{Kahng2022RTLMP}
A.~B. Kahng, R.~Varadarajan, and Z.~Wang,
``RTL-MP: Toward practical, human-quality chip
planning and macro placement'',
{\em Proc. ISPD}, 2022, pp. 3--11.

\bibitem{Kahng2024HierRTLMP}
A.~B. Kahng, R.~Varadarajan, and Z.~Wang,
``Hier-RTLMP: A hierarchical automatic macro placer
for large-scale complex IP blocks'',
{\em IEEE TCAD}, 43(5) (2024), pp. 1552--1565.

\bibitem{Kim2012}
M.-C. Kim, N. Viswanathan, C.~J. Alpert, I.~L. Markov
and S. Ramji,
``MAPLE: Multilevel adaptive placement for mixed-size designs'',
{\em Proc. ISPD}, 2012, pp. 193--200.

\bibitem{KimLim2008}
D.~H. Kim and S.~K. Lim,
``Bus-aware microarchitectural floorplanning'',
{\em Proc. ASP-DAC}, 2008, pp. 204--208.

\bibitem{Kirkpatrick1983}
S.~Kirkpatrick, C.~D. Gelatt,
and M.~P. Vecchi,
``Optimization by simulated annealing'',
{\em Science}, 220(4598) (1983), pp. 671--680.

\bibitem{Lin2019TVLSI}
J.-M. Lin, Y.-L. Deng, S.-T. Li, B.-H. Yu, L.-Y. Chang
and T.-W. Peng,
``Regularity-aware routability-driven macro placement methodology
for mixed-size circuits with obstacles'',
{\em IEEE \textcolor{black}{TVLSI}},
27(1) (2019), pp. 57--68.

\bibitem{Lin2021}
J.-M. Lin, Y.-L. Deng, Y.-C. Yang, J.-J. Chen
and P.-C. Lu,
``Dataflow-aware macro placement based on simulated evolution
algorithm for mixed-size designs'',
{\em IEEE \textcolor{black}{TVLSI}},
29(5) (2021), pp. 973--984.

\bibitem{Lu2015}
J. Lu, H. Zhuang, P. Chen, H. Chang, C.-C. Chang,
Y.-C. Wong et al.,
``ePlace-MS: Electrostatics-based placement for mixed-size
circuits'',
{\em IEEE TCAD}, 34(5) (2015), pp. 685--698.

\bibitem{Mirhoseini2020}
A. Mirhoseini, A. Goldie, M. Yazgan, J. Jiang, E. Songhori,
S. Wang et al.,
``Chip placement with deep reinforcement learning'',
{\em arXiv:2004.10746}, 2020.

\bibitem{Mirhoseini2021}
A. Mirhoseini, A. Goldie, M. Yazgan, J.~W. Jiang, E. Songhori,
S. Wang et al.,
``A graph placement methodology for fast chip design'',
{\em Nature}, 594 (2021), pp. 207--212.

\bibitem{Murata1996}
H. Murata, K. Fujiyoshi, S. Nakatake and Y. Kajitani,
``VLSI module placement based on rectangle-packing by the
sequence-pair'',
{\em IEEE TCAD}, 15(12) (1996), pp. 1518--1524.

\bibitem{Nookala2005}
V. Nookala, Y. Chen, D.~J. Lilja and S.~S. Sapatnekar,
``Microarchitecture-aware floorplanning using a statistical
design of experiments approach'',
{\em Proc. DAC}, 2005, pp. 579--584.

\bibitem{Yan2008}
J.~Z. Yan and C.~Chu,
``DeFer: Deferred decision making enabled
fixed-outline floorplanner'',
{\em Proc. DAC}, 2008, pp. 161--166.

\bibitem{Yan2014}
J.~Z. Yan, N.~Viswanathan, and C.~Chu,
``An effective floorplan-guided placement algorithm
for large-scale mixed-size design'',
{\em ACM TODAES}, 19(3) (2014), pp. 1--25.

\bibitem{ChangCC17}
C.-H. Chang, Y.-W. Chang, and T.-C. Chen, 
``A novel damped-wave
framework for macro placement'',
{\em Proc. ICCAD}, 2017, pp. 504–511.


\bibitem{Liu2024ChipNeMo}
M. Liu, T.-D. Ene, R. Kirby, C. Cheng, N. Pinckney,
R. Liang et al.,
``ChipNeMo: Domain-adapted LLMs for chip design'',
{\em arXiv:2311.00176}, 2023.

\bibitem{He2024ChatEDA}
Z. He, H. Wu, X. Zhang, X. Yao, S. Zheng, H. Zheng,
et al.,
``ChatEDA: A large language model powered autonomous agent for
EDA'',
{\em IEEE TCAD}, 43(10) (2024), pp. 3184--3197.

\bibitem{Thakur2024VeriGen}
S. Thakur, B. Ahmad, H. Pearce, B. Tan, B. Dolan-Gavitt, R. Karri et al.,
``VeriGen: A large language model for Verilog code generation'',
{\em ACM TODAES},
29(3) (2024), pp. 1--31.

\bibitem{VidalCPGM19}
 A. Vidal-Obiols, J. Cortadella, J. Petit, M. Galceran-Oms and
F. Martorell, 
``RTL-aware dataflow-driven macro placement'',
{\em Proc. DATE}, 2019, pp. 186–191.

\bibitem{UchenduGHSLJ26}
I. Uchendu, S. Goel, K. Hou, E. Songhori, K.-H. Lee, J.~W. Jiang et al.,
``See it to place it: Evolving macro placements with
vision-language models'',
{\em arXiv:2603.28733}, 2026.

\bibitem{ariane}
Ariane RISC-V CPU repo. Accessed: March 11, 2026. [Online]. \\
Available: \url{https://github.com/openhwgroup/cva6}

\bibitem{blackparrot}
BlackParrot repo. Accessed: March 11, 2026. [Online]. \\
Available: \url{https://github.com/black-parrot/black-parrot}

\bibitem{claude}
Anthropic's Claude Opus 4.6. Accessed: March 11, 2026.
[Online]. \\
Available: \url{https://www.anthropic.com/news/claude-opus-4-6}

\bibitem{hier-rtlmp}
Hier-RTLMP repo. Accessed: March 11, 2026. [Online].
Available: \\
\url{https://github.com/The-OpenROAD-Project/OpenROAD/tree/master/src/mpl}

\bibitem{innovus}
Cadence Innovus v21.1. 
\url{https://cadence.com}

\bibitem{TILOS-repo}
MacroPlacement repo. Accessed: March 11, 2026. [Online]. \\
Available: \url{https://github.com/TILOS-AI-Institute/MacroPlacement}

\bibitem{mempool}
MemPool repo. Accessed: March 11, 2026. [Online]. \\
Available: \url{https://github.com/pulp-platform/mempool}

\bibitem{ng45}
NanGate45 PDK. Accessed: March 11, 2026. [Online]. \\
Available: \url{https://eda.ncsu.edu/freepdk/freepdk45/}

\bibitem{mage_repo}
MAGE repository. [Online]. \\
Available: \url{https://github.com/ABKGroup/MAGE}

\end{thebibliography}
\end{document}